\documentclass[lettersize,journal]{IEEEtran}
\usepackage{url}
\usepackage[utf8]{inputenc} 
\usepackage[T1]{fontenc}    
\usepackage{url}            
\usepackage{booktabs}       
\usepackage{amsfonts}       
\usepackage{nicefrac}       
\usepackage{microtype}      
\usepackage{xcolor}         
\usepackage{algorithm}
\usepackage{algpseudocode}
\usepackage{times}
\usepackage{epsfig}
\usepackage{epigraph}
\usepackage{graphicx}
\usepackage{amsmath}
\usepackage{amssymb}
\usepackage{tabularx}
\usepackage{multirow}
\usepackage{float}
\usepackage{soul}
\usepackage{pifont}
\usepackage{wrapfig}
\usepackage{threeparttable}
\usepackage{mathtools}
\usepackage{xspace}
\usepackage{enumitem}
\usepackage{caption}

\usepackage{amsmath,amsfonts,bm}

\def\eqref#1{equation~\ref{#1}}

\def\1{\bm{1}}

\DeclareMathAlphabet{\mathsfit}{\encodingdefault}{\sfdefault}{m}{sl}
\SetMathAlphabet{\mathsfit}{bold}{\encodingdefault}{\sfdefault}{bx}{n}

\newcommand{\cmark}{\ding{51}}%

\newcommand{\boldparagraph}[1]{\vspace{0.2cm}\noindent{\bf #1}}

\ifCLASSOPTIONcompsoc
  \usepackage[nocompress]{cite}
\else
  \usepackage{cite}
\fi
\ifCLASSINFOpdf
\else
\fi
\def\eg{\emph{e.g.}}

\def\etc{\emph{etc}}

\begin{document}
%
\title{Senna: Bridging Large Vision-Language Models \\ and End-to-End Autonomous Driving}
%
%
\author{Bo Jiang, Shaoyu Chen, Bencheng Liao, Xingyu Zhang, Wei Yin, Qian Zhang, \\ Chang Huang, Wenyu Liu,~\IEEEmembership{Senior Member,~IEEE} and Xinggang Wang,~\IEEEmembership{Senior Member,~IEEE}
\IEEEcompsocitemizethanks{
\IEEEcompsocthanksitem B. Jiang, S. Chen, B. Liao, W. Liu, and X. Wang are with Huazhong University of Science and Technology, Wuhan 430074, China.
\IEEEcompsocthanksitem X. Zhang, W. Yin, Q. Zhang, and C. Huang are with Horizon Robotics.
\IEEEcompsocthanksitem Corresponding author: Xinggang Wang (xgwang@hust.edu.cn).
}
}

\markboth{Journal of \LaTeX\ Class Files,~Vol.~14, No.~8, October~2024}%
{Shell \MakeLowercase{\textit{et al.}}: Bare Demo of IEEEtran.cls for Computer Society Journals}

\makeatletter
\g@addto@macro\@maketitle
{
\begin{figure}[H]
\setlength{\linewidth}{\textwidth}
\setlength{\hsize}{\textwidth}
\vspace{-8pt}
\centering
\setcounter{figure}{0}
\includegraphics[width=\linewidth]{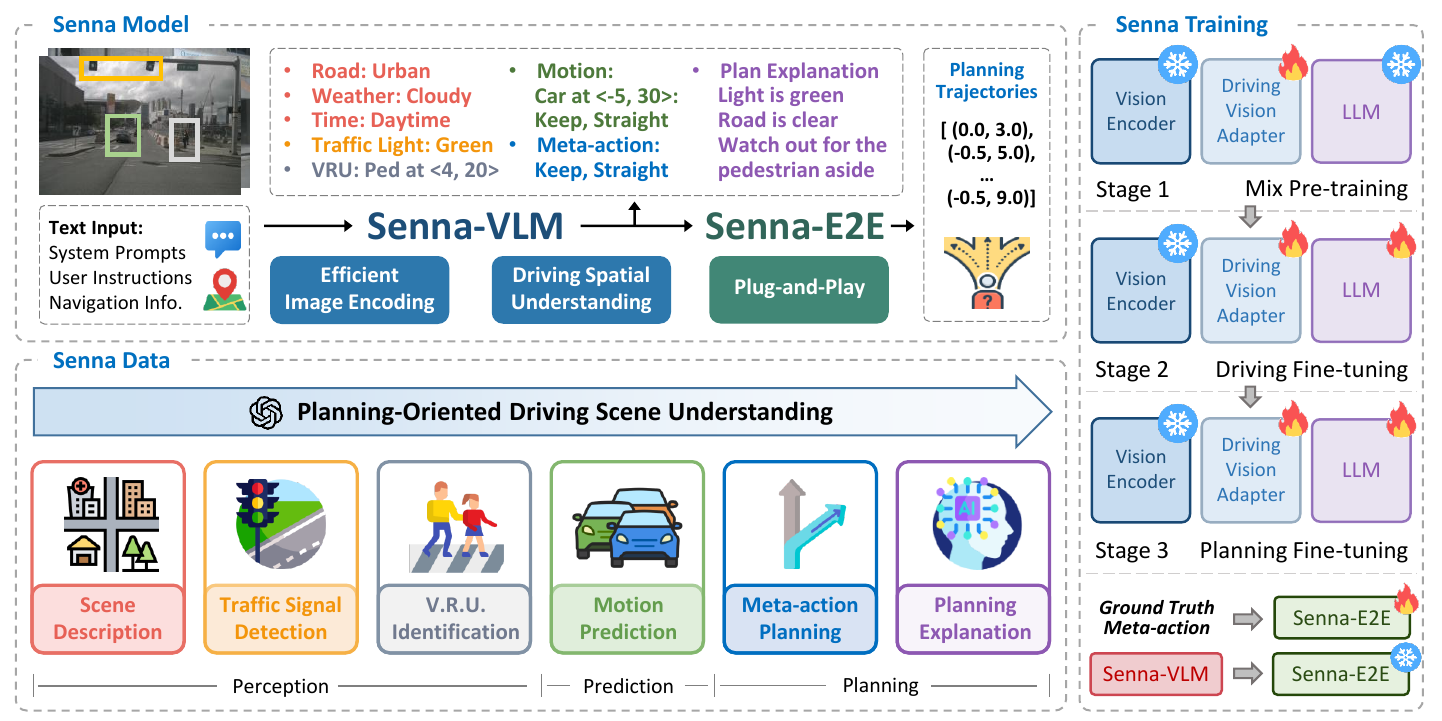}
\vspace{-8pt}
\caption{Senna is a structured autonomous driving system comprising a large vision-language model (Senna-VLM) and an end-to-end driving model (Senna-E2E). Senna-VLM generates high-level planning meta-actions in natural language, while Senna-E2E predicts low-level planning trajectories. We design a series of planning-oriented QAs that can be auto-labeled at scale, accompanied by a three-stage training strategy, to enable profound driving scene understanding and accurate planning.}
\label{fig: teaser}
\end{figure}
\vspace{-28pt}
}
\makeatother

\maketitle

\IEEEtitleabstractindextext{%
\begin{abstract}
End-to-end autonomous driving demonstrates strong planning capabilities with large-scale data but still struggles in complex, rare scenarios due to limited commonsense. In contrast, Large Vision-Language Models (LVLMs) excel in scene understanding and reasoning. The path forward lies in merging the strengths of both approaches. Previous methods using LVLMs to predict trajectories or control signals yield suboptimal results, as LVLMs are not well-suited for precise numerical predictions. This paper presents Senna, an autonomous driving system combining an LVLM (Senna-VLM) with an end-to-end model (Senna-E2E). Senna decouples high-level planning from low-level trajectory prediction. Senna-VLM generates planning decisions in natural language, while Senna-E2E predicts precise trajectories. Senna-VLM utilizes a multi-image encoding approach and multi-view prompts for efficient scene understanding. Besides, we introduce planning-oriented QAs alongside a three-stage training strategy, which enhances Senna-VLM's planning performance while preserving commonsense. Extensive experiments on two datasets show that Senna achieves state-of-the-art planning performance. Notably, with pre-training on a large-scale dataset DriveX and fine-tuning on nuScenes, Senna significantly reduces average planning error by 27.12\% and collision rate by 33.33\% over model without pre-training. We believe Senna's cross-scenario generalization and transferability are essential for achieving fully autonomous driving. Code and models will be released at \url{https://github.com/hustvl/Senna}.
\end{abstract}

\begin{IEEEkeywords}
Autonomous Driving, End-to-End, Large Vision-Language Models, Planning, Multi-Modality
\end{IEEEkeywords}}

\IEEEdisplaynontitleabstractindextext

%
\IEEEpeerreviewmaketitle

\begin{figure}[ht]
\centering
\includegraphics[width=0.98\linewidth]{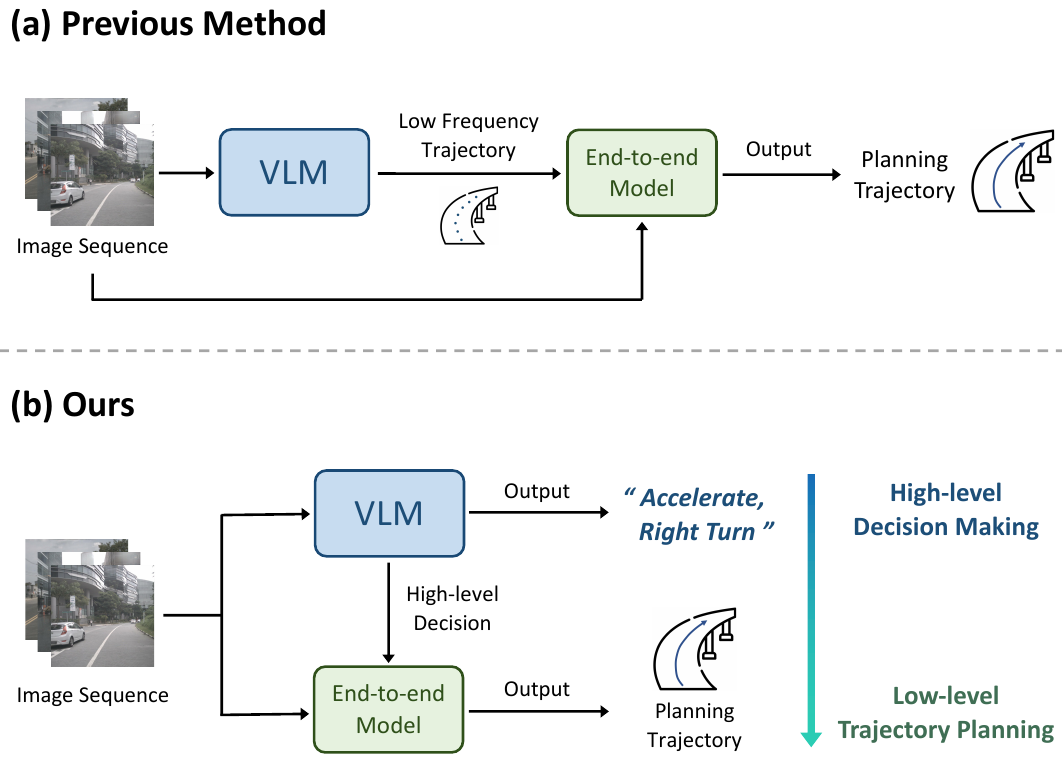} 
\caption{
Previous methods plan trajectories without a decision-making step, making model learning difficult. LVLMs also struggle with precise trajectory prediction. Senna adopts a structured planning approach: Senna-VLM leverages pre-trained commonsense and driving knowledge for high-level decisions in natural language, which Senna-E2E then uses to generate the final trajectory.
}
\label{fig:intro}
\end{figure}

\section{Introduction}
\label{sec:introduction}

\IEEEPARstart{A}{\lowercase{utonomous}} driving has witnessed rapid development in recent years~\cite{wang2022detr3d, uniad, vad}. Significant progress has been made in the areas of driving perception~\cite{philion2020lss, li2022bevformer, liao2022maptr}, motion prediction~\cite{chai2019multipath, gu2022vip3d, jiang2022pip}, and planning~\cite{toromanoff2020end, transfuser, hu2022stp3}. These developments provide a solid foundation for achieving more accurate and safer driving decisions. Among them, end-to-end autonomous driving stands out as a significant breakthrough. Powered by large-scale data, end-to-end approaches have demonstrated remarkable planning capabilities. Besides, Large vision-language models (LVLMs)~\cite{liu2024llava, bai2023qwenvl, wang2023cogvlm, chen2024internvl, achiam2023gpt4} have demonstrated increasingly powerful image understanding and reasoning abilities. By leveraging their common sense and logic, LVLMs can analyze driving environments and make safe decisions in complex scenarios. Utilizing vast driving data to improve the performance of LVLMs in autonomous driving and bridging LVLMs and end-to-end models is paramount to achieving safe, robust, and generalizable autonomous driving.

The common practice in end-to-end autonomous driving is to directly predict future trajectories or control signals without a decision-making step. However, this approach may make model learning more difficult while lacking interpretability. In contrast, when the human brain makes detailed decisions, a system composed of hierarchical high-level decision-making and low-level execution plays a crucial role~\cite{botvinick2008hierarchical, koechlin2003cognitive, badre2008cognitive}. Additionally, end-to-end models often lack common sense and can make errors in simple scenarios. For example, they may misinterpret a truck carrying traffic cones as a roadblock, triggering unnecessary braking. These limitations hinder the planning performance of end-to-end models.

In this paper, we primarily explore and attempt to answer the following three key questions.

\boldparagraph{How to integrate LVLMs with end-to-end models?}
The application of LVLMs in autonomous driving planning is currently categorized into two main types. One is using LVLMs directly as planners to predict trajectory points or control signals~\cite{drivegpt4, driving-with-llms, drivemlm, wang2024omnidrive}. However, LVLMs inherently have limitations in precise mathematical computations~\cite{frieder2024mathematical, hendrycks2021measuring}, making them less suitable for predicting accurate numbers. The other approach is to combine LVLMs with end-to-end models. Previous pioneering work~\cite{tian2024drivevlm} involves using LVLMs to predict low-frequency trajectory points, which are then refined by the end-to-end model to produce high-frequency trajectories. This reduces the number of points predicted by LVLMs, partially mitigating the issue. However, since LVLMs are still required to predict trajectory points, the problem is not entirely resolved.

In this work, we propose Senna, a structured autonomous driving system that integrates a Large Vision-Language Model (Senna-VLM) with an end-to-end model (Senna-E2E), named after the renowned Brazilian racing driver Ayrton Senna. Senna's structured planning philosophy lies in its decoupling of high-level planning decisions and low-level trajectory predictions. Specifically, Senna-VLM predicts high-level planning decisions in natural language and encodes them into high-dimensional features, which are then fed into Senna-E2E. Conditioning on the high-level decisions, Senna-E2E generates the final planning trajectories.

Predicting language-based planning decisions allows LVLMs to fully utilize their knowledge and common sense from pre-trained language tasks, so as to generate reasonable decisions and avoid issues with precise numerical predictions. End-to-end models, in contrast, excel at accurate trajectory prediction. By decoupling high-level decision-making and low-level trajectory planning, we reduce the learning complexity for the end-to-end model and improve the trajectory planning accuracy.

Similar to previous work~\cite{drivemlm, tian2024drivevlm}, Senna-VLM uses formatted meta-actions for planning decisions to narrow the prediction space, which reduces the learning difficulty for the LVLM and makes it easier for the integration with the end-to-end model. A straightforward meta-action encoder is utilized to convert planning decisions into high-dimensional features, which are then fed into Senna-E2E. During training, Senna-E2E uses ground truth planning decisions as input, whereas, during inference, it relies on decisions predicted by Senna-VLM.

\boldparagraph{How to design an LVLM tailored for driving tasks?}
Current popular LVLMs are not specifically optimized for multi-image input~\cite{liu2024llava, bai2023qwenvl, lin2024vila}. Previous LVLMs for driving tasks either support only front-view input~\cite{drivegpt4, zhou2024elm}, which limits spatial awareness and increases safety risks, or they accommodate multi-image input but still lack a detailed design or validation of their effectiveness~\cite{drivemlm, tian2024drivevlm}.

Senna-VLM is tailored for driving tasks. It supports multi-image inputs to encode surround-view data, which is essential for understanding driving scenarios and ensuring safety. Initially, we attempted to integrate surround-view input using LLaVA-1.5~\cite{liu2024llava1.5}, but the results were unsatisfactory due to excessive image tokens. To address this, we propose a driving vision adapter to encode and compress image tokens and introduce prompts specifically designed for surround-view scenarios. These designs help the model differentiate between image features from various perspectives and develop spatial understanding.

\boldparagraph{How to effectively train a driving LVLM?}
After developing an LVLM for driving tasks, the final step is ensuring effective training, which requires both suitable data and strategies.  In terms of data, while previous works have proposed various data strategies~\cite{sima2023drivelm, wu2023nuprompt, qian2024nuscenes-qa}, many are not specifically designed for planning tasks, but for other tasks like detection and grounding. In this paper, we are the first to validate the significance of different types of QAs in driving planning. Specifically, we introduce a series of planning-oriented QAs designed to enhance the VLM's understanding of planning-related clues in driving scenarios, ultimately enabling more accurate planning. These QAs include driving scene descriptions, motion intention prediction of traffic participants, traffic signal detection, meta-action planning, \etc. Unlike previous works~\cite{tian2024drivevlm, zhou2024elm} that relied on manual annotation, our data strategy can be produced at scale entirely through an automatic pipeline.

As for training strategies, most existing approaches adopt a general-purpose pre-training followed by driving fine-tuning~\cite{drivemlm, tian2024drivevlm}. However, our experimental results indicate that this may not be the optimal choice. Instead, we propose a three-stage training strategy for Senna-VLM, consisting of mixed pre-training, driving fine-tuning, and planning fine-tuning. Experimental results demonstrate that our proposed training strategy leads to the best planning performance.

Senna offers several advantages. Its structured planning strategy combines the strengths of LVLMs and end-to-end models to improve the safety, robustness and generalization of autonomous driving, while its support for surround-view and multi-image input effectively enhances its ability for driving scene understanding. The proposed planning-oriented QAs and the three-stage training strategy enable Senna-VLM to make more accurate planning decisions while preserving its common sense, preventing model collapse.

Our main contributions can be summarized as follows:
\begin{itemize}
\item We propose Senna, an autonomous driving system that integrates an LVLM with an end-to-end model, achieving structured planning from high-level decisions to low-level trajectory prediction. Senna-VLM utilizes common sense and logical reasoning to analyze the scene and outputs decisions in natural language, Senna-E2E then generates concrete planning trajectories based on the decisions.
\item Senna employs an efficient multi-image encoding strategy along with a carefully designed surround-view prompt, enabling accurate perception and spatial understanding of driving scenarios.
\item  We design a series of planning-oriented QAs and a three-stage training strategy to enhance Senna's driving understanding and planning performance while preserving its common-sense knowledge.
\item Extensive experiments from the nuScenes~\cite{caesar2020nuscenes} dataset and a large-scale dataset DriveX demonstrate Senna's SOTA planning performance. Notably, by leveraging pre-trained weights from the DriveX dataset and fine-tuning on the nuScenes dataset, Senna achieves significant performance gains, demonstrating strong cross-scenario generalization and transferability. These results validate the effectiveness and versatility of Senna’s structured planning approach, architecture design, and training strategy.
\end{itemize}

\begin{figure*}[ht]
\centering
\includegraphics[width=0.98\textwidth]{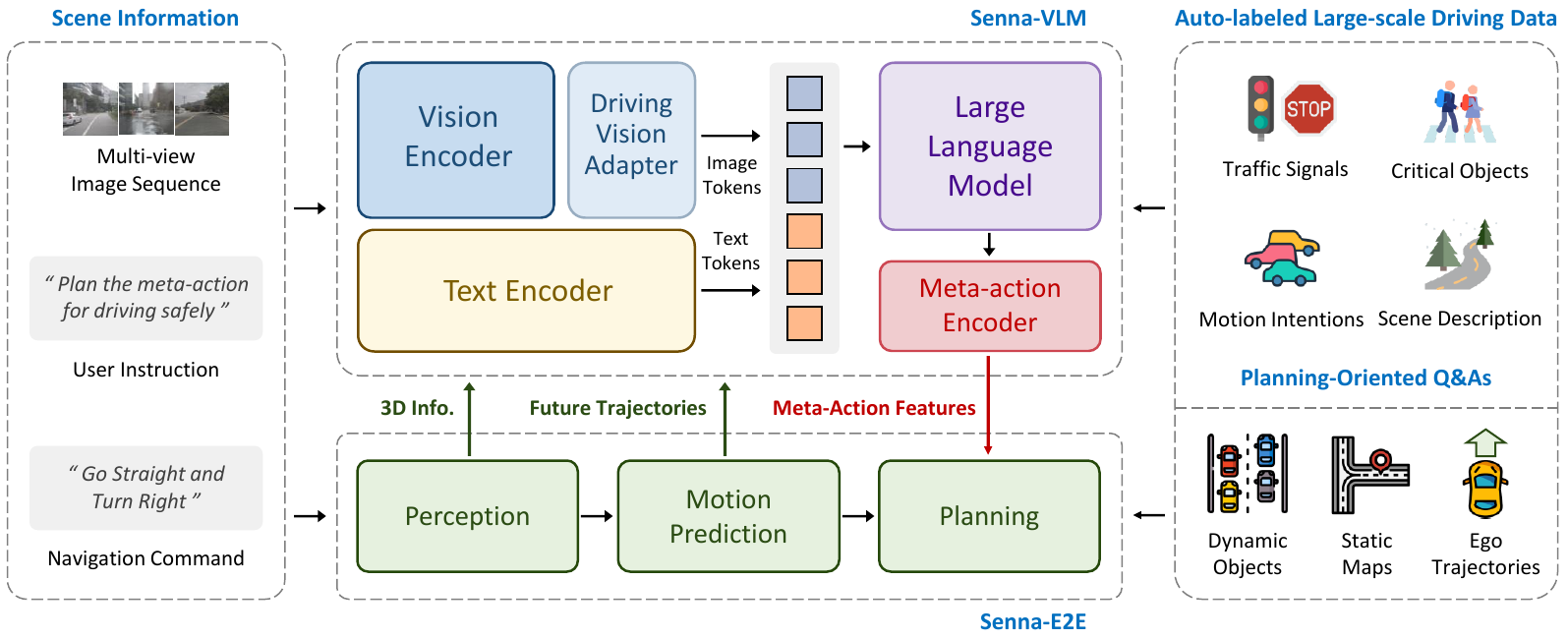} 
\caption{\textbf{Overall architecture of Senna.} Senna consists of two modules: Senna-VLM and Senna-E2E. Senna-VLM efficiently encodes multi-view images and achieves spatial understanding from a surrounding perspective. Given the input scene information, Senna-VLM predicts high-level planning decisions, which are encoded into meta-action features. Senna-E2E then predicts the final planning trajectories based on both the scene information and the meta-action features. We train Senna using auto-labeled large-scale driving data.}
\label{fig:framework}
\end{figure*}

\section{Related Work}

\boldparagraph{End-to-end Autonomous Driving.} Traditional autonomous driving systems typically employ a staged, modular design with rule-based planners~\cite{paden2016survey, thrun2006stanley, urmson2008autonomous}, which encounter challenges such as limited model generalization and a ceiling on planning performance. In contrast, end-to-end autonomous driving leverages neural networks to directly map perception inputs to planning outputs. Early works~\cite{codevilla2019exploring, pomerleau1988alvinn, transfuser} treat this as a black-box approach, and despite the lack of interpretability, they demonstrate the potential of neural networks in end-to-end learning. With the rapid advancement of autonomous driving in recent years, the explosive growth in available driving data has provided substantial support for end-to-end approaches~\cite{roach, driveadapter, maptrv2, uniad, vad, chen2024vadv2}. UniAD~\cite{uniad} introduces a multi-task framework using Transformers~\cite{vaswani2017attention} to perform object tracking, trajectory prediction, map prediction, occupancy prediction, and planning. The additional supervision from these auxiliary tasks significantly enhances the model's planning performance. VAD~\cite{vad} introduces vectorized scene representation, striking a better balance between accuracy and speed. VADv2~\cite{chen2024vadv2} further proposes probabilistic planning, replacing deterministic planning with multimodal trajectory scoring to better capture planning uncertainty. We extend VADv2 to incorporate high-level decision information, which serves as the end-to-end model for Senna.

\boldparagraph{Large Vision-Language Models.} With the emergence of large language models (LLMs) exhibiting powerful understanding and conversational capabilities~\cite{gpt3, touvron2023llama, anil2023palm2, yang2024qwen2}, it is natural to extend this capacity from a single textual modality to multimodal systems, with visual modality being one of the most critical~\cite{chen2024internvl, liu2024llava, bai2023qwenvl, lin2024vila, zhu2023minigpt4, wang2023cogvlm, dai2023instructblip}. CLIP~\cite{radford2021clip, fang2023eva} pioneered the use of image-text pairs for contrastive learning, enabling models to achieve large-scale pre-training in a self-supervised manner. BLIP~\cite{li2022blip, li2023blip2} bridges vision and language models through cross-modal contrastive learning and generative tasks. Building on these cross-modal training strategies~\cite{jia2021align, lu2019vilbert, radford2021clip, tan2019lxmert}, LVLMs have achieved a series of advancements. GPT-4V~\cite{achiam2023gpt4} demonstrates remarkable performance in understanding complex scenes and multi-task reasoning. LLaVA~\cite{liu2024llava, liu2024llava1.5} introduces the visual instruction tuning, leveraging GPT-4 generated language-image instruction-following data for cross-modal learning. QwenVL~\cite{bai2023qwenvl} makes improvements in visual connectors, training strategies, and data, excelling in multilingual understanding and 3D perception. VILA~\cite{lin2024vila} enhances multimodal performance by optimizing cross-modal pre-training and supervised fine-tuning strategies. Qwen2-VL~\cite{wang2024qwen2vl} adopts the more advanced Qwen2~\cite{yang2024qwen2} as its language model and introduces the use of multimodal rotary position embedding and a dynamic number of visual tokens to support image and video inputs of arbitrary resolutions.

\boldparagraph{Large Vision-Language Models \& Autonomous Driving.} The commonsense knowledge, reasoning abilities, and interpretability of LVLMs can effectively complement the shortcomings of end-to-end models. Drive-with-LLMs~\cite{driving-with-llms} uses ground truth driving perception data and a Transformer network to encode perception information into latent space, which is then fed into an LLM to predict future planning trajectories. DriveGPT4~\cite{drivegpt4} takes front camera video inputs and uses LVLMs to predict planning control signals and provide decision explanations. LanguageMPC~\cite{languagempc} converts historical ground truth perception information and HD maps to natural language format, and uses chain of thought reasoning to analyze driving scenes and generate planning actions. DriveMLM~\cite{drivemlm} validate the effectiveness of LVLM-based planning models in a closed-loop simulation environment~\cite{dosovitskiy2017carla}. ELM~\cite{zhou2024elm} introduces large-scale pre-training of LVLMs using internet-scale cross-domain video data, demonstrating that incorporating diverse sources and task-specific training data significantly improves the LVLM's performance in driving tasks. Several other studies proposes data collection strategies and datasets tailored for driving tasks, further advancing the LVLM development in autonomous driving~\cite{sima2023drivelm, wu2023nuprompt, qian2024nuscenes-qa, wu2023refer-kitti, deruyttere2019talk2car, kim2018bdd-x}. DriveVLM~\cite{tian2024drivevlm} is the first to combine LVLMs with end-to-end models, where LVLMs predict a low-frequency trajectory, and the end-to-end model refines it to generate the final planning trajectory. DriveVLM showcases strong planning performance on both the nuScenes dataset and their proposed dataset. However, since LVLMs are not well-suited for precise mathematical computations, using LVLMs to predict trajectory points may result in sub-optimal performance. We propose a structured planning strategy. Senna-VLM predicts high-level decision information in natural language, avoiding the need for LVLMs to predict precise numerical values. This approach aligns with the LVLM pre-training tasks that involve natural language prediction, effectively leveraging pre-trained knowledge to make accurate planning decisions. Based on Senna-VLM's predicted decision, Senna-E2E will then generate the final planning trajectory.


\section{Senna}
\label{sec:method}

In this section, we provide a detailed introduction to Senna. Fig.~\ref{fig:framework} illustrates the overall architecture of Senna. The input scene information includes multi-view image sequence, user instruction, and navigation command. The user instruction is fed into Senna-VLM as prompts, while the others are sent to both Senna-VLM and Senna-E2E. Senna-VLM encodes the image and text information into image and text tokens, respectively, which are then processed by the LLM. The LLM generates high-level decisions, which are encoded into high-dimensional features via the Meta-action Encoder. Senna-E2E predicts the final planning trajectory based on the scene information and the meta-action features generated by Senna-VLM. We design a series of planning-oriented QAs for training Senna-VLM, which do not require human annotation and can be produced at scale entirely through an auto-labeling pipeline.

In Sec~\ref{sec:qa}, we detail the planning-oriented QAs for driving scene understanding. Sec~\ref{sec:vlm} and Sec~\ref{sec:e2e} focus on the design of Senna-VLM and Senna-E2E, respectively. Sec~\ref{sec:train} introduces our proposed three-stage training strategy.

\subsection{Driving Scene Understanding}
\label{sec:qa}
\begin{figure*}[thbp]
\centering
\includegraphics[width=0.98\textwidth]{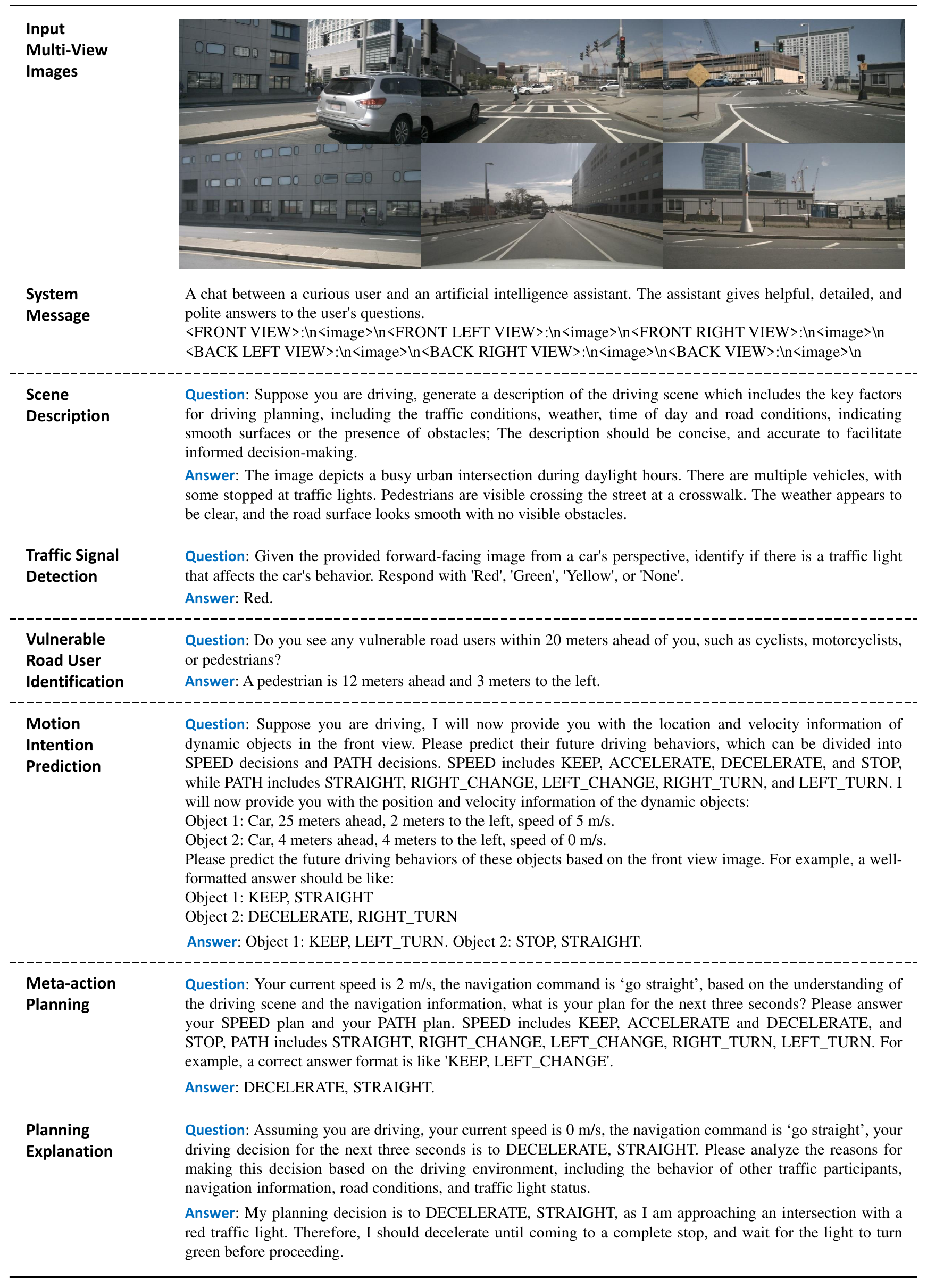}
\vspace{-8pt}
\caption{Visualization of our proposed planning-oriented QAs.}
\label{fig:conversation}
\vspace{-25pt}
\end{figure*}

Understanding key factors in driving scenes is crucial for safe and accurate planning. We design a series of planning-oriented QAs to enhance Senna-VLM's comprehension of driving scenes. The detail of each type of QA is illustrated in Fig.~\ref{fig:conversation}. The raw data used to generate these QAs, such as 3D object detection boxes and object tracking trajectories, can be obtained through automatic annotation systems. Besides, description QAs can be generated by LVLMs like GPT-4o~\cite{achiam2023gpt4}. We will elaborate on each type of QA below.

\begin{figure}[ht]
\centering
\includegraphics[width=0.98\linewidth]{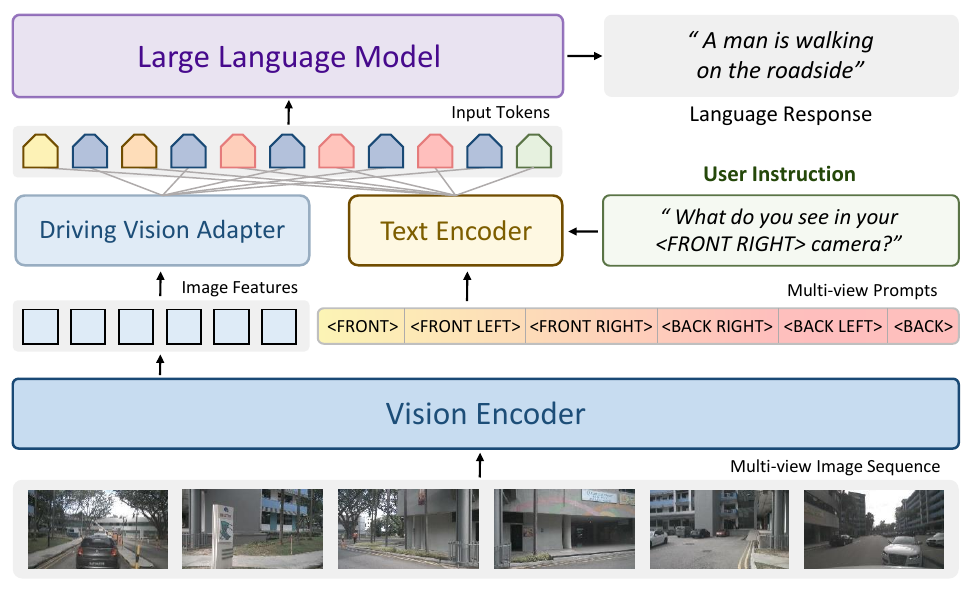} 
\caption{
Our proposed multi-view prompts and image encoding approach for efficient driving scene understanding.
}
\label{fig:adapter}
\end{figure}

\boldparagraph{Scene Description.}
We utilize a pre-trained LVLM to generate descriptions of driving scenes based on surround-view images. To avoid generating redundant information irrelevant to planning, we specify the required information in the prompts, which include: traffic conditions, the environment (\eg, urban, rural, \etc.), road types (\eg, paved roads, highways), weather conditions, time of day, and road conditions (\eg, whether the road is smooth or if there are any obstacles). By structuring the prompts this way, we can obtain concise and informative scene descriptions.

\boldparagraph{Traffic Signal Detection.} 
There are various types of traffic signals, but here we primarily focus on the most critical one: traffic lights. The states of traffic lights can be categorized into four types: red, green, yellow, and none, where none indicates that no traffic light is detected in front of the ego vehicle.

\boldparagraph{Vulnerable Road User Identification.}
By identifying vulnerable road users (VRUs) in the environment, we enhance Senna's perception of these critical objects and improve the safety of planning. Specifically, we use ground truth 3D detection results to obtain the categories and locations of VRUs and then describe this information in text form. The location information is centered on the ego vehicle, including the lateral and longitudinal distance of each VRU relative to the ego vehicle. We only use Senna-VLM to predict the integer part of the distance to reduce the learning complexity while building distance perception.

\boldparagraph{Motion Intention Prediction.}
Accurately predicting the future motion intentions of other vehicles is a prerequisite for safe planning. We also adopt a meta-action approach, allowing Senna to predict the future behavior of surrounding vehicles. This enhances Senna's understanding of the dynamic features of the scene and enables it to make more informed decisions.

\boldparagraph{Meta-action Planning.} 
To avoid using LVLMs for precise trajectory prediction, we convert the ego vehicle's future trajectory into meta-actions for high-level planning. Specifically, meta-actions consist of lateral and longitudinal decisions. Lateral meta-actions include \textit{Left}, \textit{Straight}, and \textit{Right}, while longitudinal meta-actions cover \textit{Accelerate}, \textit{Keep}, \textit{Decelerate}, and \textit{Stop}. Lateral meta-actions are determined based on the lateral displacement during the predicted future time step $T$, and longitudinal meta-actions are determined by the change of speed over the predicting period. The final meta-action consists of both lateral and longitudinal meta-actions.

\boldparagraph{Planning Explanation.}
We also generate planning explanations using LVLMs based on the vehicle's ground truth future movements. In other words, we inform LVLMs about the vehicle's actual future motions (\eg, accelerating and turning left) and ask them to analyze the reasons behind such decisions. In the prompt, we guide the model to analyze the decision by considering the following factors that influence planning: the behavior of other traffic participants, navigation information, road conditions, and traffic light status.

\begin{table*}
\small
\centering
\setlength\tabcolsep{6pt}
\begin{tabular}{l|c|ccc|cccc|ccc}
\toprule
\multirow{2}{*}{Method} & \multirow{2}{*}{Acc. $(\%)$ $\uparrow$} & \multicolumn{3}{c|}{Path $($F1$)$ $\uparrow$} & \multicolumn{4}{c|}{Speed $($F1$)$ $\uparrow$} & \multirow{2}{*}{BLEU-4 $\uparrow$} & \multirow{2}{*}{CIDEr $\uparrow$} & \multirow{2}{*}{METEOR $\uparrow$}\\
& & straight & left & right & keep & acc. & dec. & stop & & \\
\midrule
QwenVL~\cite{bai2023qwenvl} & 7.66 & 78.95 & 4.87 & 1.14 & 3.28 & 2.89 & 15.14 & 1.21 & 8.97 & 4.01 & 31.04 \\
VILA~\cite{lin2024vila} & 6.95 & 31.70 & 0.00 & 2.52 & 65.37 & 10.79 & 0.00 & 0.00 & 14.65 & 9.67 & 36.43 \\
LLaVA-1.5~\cite{liu2024llava1.5} & 21.34 & 87.18 & 0.00 & 0.00 & 64.78 & 0.00 & 0.00 & 0.00 & 14.05 & 5.71 & 36.94 \\
\midrule
QwenVL${\text{\textdagger}}$~\cite{bai2023qwenvl} & 56.38 & \textbf{96.01} & 90.48 & 89.80 & 66.49 & 49.88 & 45.22 & 64.78 & 25.89 & 19.91 & 46.54 \\
VILA${\text{\textdagger}}$~\cite{lin2024vila} & 63.03 & 95.74 & \textbf{90.74} & 88.30 & 73.65 & 43.14 & 49.81 & 77.65 & 23.50 & 17.23 & 44.17 \\
LLaVA-1.5${\text{\textdagger}}$~\cite{liu2024llava1.5} & 64.48 & 96.00 & 89.04 & 89.57 & 74.94 & 44.24 & 52.68 & 78.96 & 25.97 & 21.62 & 45.08 \\
Senna & \textbf{71.21} & 95.60 & 89.37 & \textbf{90.09} & \textbf{80.18} & \textbf{58.83} & \textbf{61.99} & \textbf{80.10} & \textbf{31.04} & \textbf{34.31} & \textbf{50.44} \\
\bottomrule
\end{tabular}
\caption{\textbf{High-level planning and scene description evaluation results on the DriveX validation dataset.} Senna demonstrates a deeper understanding of driving scenarios compared to previous approaches, leading to more accurate planning decisions. ${\text{\textdagger}}$ denotes fine-tuned on the DriveX dataset. "acc." refers to accelerate and "dec." refers to decelerate, respectively.}
\label{tab:main}
\end{table*}

\subsection{Senna-VLM}
\label{sec:vlm}
Senna-VLM consists of four components. The Vision Encoder takes multi-view image sequences $I \in \mathbb{R}^{N_{\rm img} \times H \times W \times 3}$ as input and extracts image features, which are further encoded and compressed by the Driving Vision Adapter, resulting in image tokens $E_{\rm img} \in \mathbb{R}^{N_{\rm img} \times M_{\rm img} \times C}$. $N_{\rm img}$, $M_{\rm img}$, and $C$ is the number of images, the number of image tokens per image, and the feature dimension of the LLM, respectively. $H$ and $W$ are the height and width of the image respectively. The Text Encoder encodes user instructions and navigation commands into text tokens $E_{\rm txt} \in \mathbb{R}^{M_{\rm txt} \times C}$, where $M_{\rm txt}$ is the number of text tokens. Both the image and the text tokens are fed into the LLM, which predicts high-level decisions. In practice, Vicuna-v1.5-7b~\cite{zheng2023vicuna} is used as our LLM. Finally, the Meta-action Encoder encodes the decisions and outputs meta-action features $e_{\rm act} \in \mathbb{R}^{D}$, where $D$ is the feature dimension of Senna-E2E.

We use ViT-L/14 from CLIP~\cite{radford2021clip} as the Vision Encoder, where each image is resized to $H=W=224$, resulting in 576 image tokens. With multi-image inputs, this leads to an excessive number of image tokens, which not only slows down the VLM training and inference but also causes model collapse and decoding failure. Therefore, we introduce a Driving Vision Adapter module. this module not only functions similarly to previous studies~\cite{li2023blip2, liu2024llava} which maps image features to the LLM feature space but also performs additional encoding and compression of the image features to reduce the number of image tokens. Specifically, we adopt a group of image queries $Q_{img} \in \mathbb{R}^{N_{\rm img} \times M_{\rm img} \times C}$ to encode the image features and output image tokens:

\begin{equation}
E_{\rm img}^{i} = \texttt{MHSA}[Q_{\rm img}^{i}, \texttt{W} \cdot \textit{g}(I^{i}), \texttt{W} \cdot \textit{g}(I^{i})] \in \mathbb{R}^{M_{\rm img} \times C},
\label{eq:adapter}
\end{equation}

\noindent where \texttt{MHSA} refers to multi-head self attention~\cite{vaswani2017attention}, \textit{g} is the vision encoder and \texttt{W} is a multi-layer perceptron composed of linear layers and GELU~\cite{hendrycks2016gelu} activation layers to project image features. $E_{\rm img}^{i}$ and $Q_{\rm img}^{i}$ denotes image tokens and image queries for the $i$-th image $I^{i}$, respectively. Our experiments demonstrate that further encoding and compressing image features does not degrade model performance. However, an excessive number of image tokens will lead to model collapse and decoding failure.

\begin{table}[]
\begin{center}
\centering
\renewcommand{\tabcolsep}{4.5pt}
\begin{tabular}{l|cccc|cccc}
\toprule
\multirow{2}{*}{Method} &
\multicolumn{4}{c|}{L2 (m) $\downarrow$} & 
\multicolumn{4}{c}{Collision (\%) $\downarrow$} \\
& 1s & 2s & 3s & Avg. & 1s & 2s & 3s & Avg. \\
\midrule
IL~\cite{ratliff2006il} & 0.44 & 1.15 & 2.47 & 1.35 & 0.08 & 0.27 & 1.95 & 0.77 \\
NMP~\cite{zeng2019nmp} & - & - & - & 2.31 & - & - & 1.92 & - \\
SA-NMP~\cite{zeng2019nmp} & - & - & - & 2.05 & - & - & 1.59 & - \\
FF~\cite{hu2021ff} & 0.55 & 1.20 & 2.54 & 1.43 & 0.06 & 0.17 & 1.07 & 0.43 \\
EO~\cite{khurana2022eo} & 0.67 & 1.36 & 2.78 & 1.60 & 0.04 & 0.09 & 0.88 & 0.33 \\
ST-P3~\cite{hu2022stp3} & 1.33 & 2.11 & 2.90 & 2.11 & 0.23 & 0.62 & 1.27 & 0.71 \\
UniAD~\cite{uniad} & 0.48 & 0.96 & 1.65 & 1.03 & 0.05 & 0.17 & 0.71 & 0.31 \\
\midrule
VAD-Base*~\cite{vad} & 0.17 & 0.34 & 0.60 & 0.37 & 0.07 & 0.10 & 0.24 & 0.14 \\
DriveVLM*~\cite{tian2024drivevlm} & 0.15 & 0.29 & 0.48 & 0.31 & 0.05 & \textbf{0.08} & 0.17 & 0.10 \\
Senna* & \textbf{0.11} & \textbf{0.21} & \textbf{0.35} & \textbf{0.22} & \textbf{0.04} & \textbf{0.08} & \textbf{0.13} & \textbf{0.08} \\
\midrule
VAD-Base~\cite{vad} & 0.41 & 0.70 & 1.05 & 0.72 & 0.07 & 0.17 & 0.41 & 0.22 \\
Senna & 0.37 & 0.54 & 0.86 & 0.59 & 0.09 & 0.12 & 0.33 & 0.18 \\
Senna$\dagger$ & \textbf{0.26} & \textbf{0.42} & \textbf{0.61} & \textbf{0.43} & \textbf{0.05} & \textbf{0.11} & \textbf{0.21} & \textbf{0.12} \\
\bottomrule
\end{tabular}
\end{center}
\caption{\textbf{Trajectory planning results on the nuScenes validation dataset.} * denotes using ego status features as input. In the experiments on the nuScenes dataset, we incorporate Senna with VAD~\cite{vad} for a fair comparison. $\dagger$ denotes initialization using pre-trained weights on the DriveX dataset.}
\label{tab:traj_nusc}
\end{table}

To enable Senna-VLM to distinguish image features from different views and build spatial understanding, we design a simple yet effective surround-view prompt for driving scenarios. Taking the front view as an example, the corresponding prompt is: \verb|<FRONT VIEW>:\n<image>\n|, where \verb|<image>| is a special token for the LLM and will be replaced by the image tokens during generation. Fig.~\ref{fig:adapter} illustrates the design of our proposed multi-view prompts and image encoding approach.

Finally, we propose the Meta-action Encoder $\varphi$, which transforms the high-level decisions output by the LLM into meta-action features $e_{\rm act}$. Since we use a formatted set of meta-actions, the Meta-action Encoder $\varphi$ achieves one-to-one mapping from meta-actions to meta-action features using a group of learnable embeddings $E_{\rm act} \in \mathbb{R}^{N_{\rm act} \times D}$, where $N_{\rm act}$ is the number of meta-actions. Eq.~\ref{eq:meta-action} illustrates the process of generating meta-action features:

\begin{equation}
e_{\rm act} = \varphi(\texttt{LLM}[\texttt{concat}(\{E_{\rm view}^{i}, E_{\rm img}^{i}\}_{i=1}^{N_{\rm img}}, E_{\rm txt})]),
\label{eq:meta-action}
\end{equation}

\noindent where $E_{\rm view}^{i}$ is the surround-view prompt corresponding to the $i$-th image. \texttt{LLM} is the large language model of Senna. Subsequently, the meta-action features will be fed into Senna-E2E to predict the planning trajectory.

\subsection{Senna-E2E}
\label{sec:e2e}
Senna-E2E extends VADv2~\cite{chen2024vadv2}. Specifically, the input to Senna-E2E includes a multi-view image sequence, the navigation command, and the meta-action features. It consists of three modules: the perception module, which detects dynamic objects and generates a local map; the motion prediction module, which predicts the future trajectories of dynamic objects; and the planning module, which uses a group of planning tokens that interact with scene features via the attention mechanism~\cite{vaswani2017attention} to predict the planning trajectory $V$. We integrate meta-action features as an additional interaction token for Senna-E2E. Since meta-action features are in the form of embedding vectors, Senna-VLM can easily be combined with other end-to-end models. The trajectory planning process of Senna-E2E $\Phi$ can be formulated as follows:

\begin{equation}
V = \Phi(I, e_{\rm nav}, e_{\rm act}) \in \mathbb{R}^{T \times 2},
\end{equation}

\noindent where $e_{\rm nav}$ is navigation commands.

\begin{table}[]
\begin{center}
\centering
\renewcommand{\tabcolsep}{4.0pt}
\resizebox{0.95\linewidth}{!}{
\begin{tabular}{l|c|ccc}
\toprule
\multirow{2}{*}{Method} & Planning & \multicolumn{3}{c}{L2 (m) $\downarrow$} \\
 & Information & 1s & 2s & 3s  \\
\midrule
VADv2~\cite{chen2024vadv2} & G.T. Meta-action & 0.37 & 1.08 & 2.24 \\
\midrule
VADv2~\cite{chen2024vadv2} & None & 0.55 & 1.46 & 3.10 \\
DriveVLM$\dagger$~\cite{tian2024drivevlm} & Low Freq. Traj. & 0.61 & 1.44 & 3.00 \\
Senna & Pred. Meta-action & \textbf{0.49} & \textbf{1.27} & \textbf{2.62} \\
\bottomrule
\end{tabular}
}
\end{center}
\caption{\textbf{Trajectory planning results on the DriveX validation dataset.} $\dagger$: our reproduced DriveVLM~\cite{tian2024drivevlm}, which employs the same vision encoder and LLM as Senna. Low Freq. Traj. denotes Low Frequency Trajectory.}
\label{tab:traj_drivex}
\end{table}

\begin{table*}[]
\begin{center}
\centering
\resizebox{0.98\textwidth}{!}{
\begin{tabular}{c|cccccc|c|ccc|cccc}
\toprule
\multirow{2}{*}{ID} & Meta & Scene & Traffic & Plan & VRU & Mot. & \multirow{2}{*}{Acc. $(\%)$ $\uparrow$} & \multicolumn{3}{c|}{Path $($F1$)$ $\uparrow$} & \multicolumn{4}{c}{Speed $($F1$)$ $\uparrow$} \\
& Action & Desc. & Light & Expl. & Idet. & Pred. & & straight & left & right & keep & acc. & dec. & stop  \\
\midrule
1 &  & \cmark & \cmark & \cmark & \cmark & \cmark & 52.67 & 90.69 & 77.08 & 69.97 & 72.24 & 9.26 & 13.78 & 69.42 \\
2 & \cmark &  & \cmark & \cmark & \cmark & \cmark & 53.65 & 94.04 & 89.13 & 83.38 & 58.83 & 40.59 & 34.81 & 79.30 \\
3 & \cmark & \cmark &  & \cmark & \cmark & \cmark & 57.39 & 94.61 & 88.27 & 88.37 & 66.13 & 43.52 & 37.47 & 78.87 \\
4 & \cmark & \cmark & \cmark &  & \cmark & \cmark & 67.91 & 95.12 & 89.32 & 87.95 & 77.70 & 51.20 & 52.57 & 78.75 \\
5 & \cmark & \cmark & \cmark & \cmark & & \cmark & 68.44 & 95.43 & \textbf{90.75} & \textbf{91.17} & 78.24 & 53.80 & 53.36 & 79.72 \\
6 & \cmark & \cmark & \cmark & \cmark & \cmark & & 68.70 & 95.12 & 89.11 & 86.88 & 78.40 & 52.99 & 51.60 & 78.53 \\
7 & \cmark & \cmark & \cmark & \cmark & \cmark & \cmark & \textbf{71.21} & \textbf{95.60} & 89.37 & 90.09 & \textbf{80.18} & \textbf{58.83} & \textbf{61.99} & \textbf{80.10} \\
\bottomrule
\end{tabular}
}
\end{center}
\caption{\textbf{Ablations of the effectiveness of planning-oriented QAs.} "Desc." denotes Description. "Expl." denotes Explanation. "VRU Idet." denotes Vulnerable Road User Identification. "Mot. Pred." indicates Motion Intention Prediction.}
\label{tab:qa}
\end{table*}

\begin{table}[]
\begin{center}
\centering
\renewcommand{\tabcolsep}{3.0pt}
\resizebox{0.98\linewidth}{!}{
\begin{tabular}{c|c|ccc|cccc}
\toprule
Image & \multirow{2}{*}{Acc. $(\%)$ $\uparrow$} & \multicolumn{3}{c|}{Path $($F1$)$ $\uparrow$} & \multicolumn{4}{c}{Speed $($F1$)$ $\uparrow$} \\
View & & stra. & left & right & keep & acc. & dec. & stop  \\
\midrule
Front & 64.91 & 94.16 & 86.90 & 83.27 & 76.24 & 51.78 & 50.53 & 79.74 \\
Surround & \textbf{71.21} & \textbf{95.60} & \textbf{89.37} & \textbf{90.09} & \textbf{80.18} & \textbf{58.83} & \textbf{61.99} & \textbf{80.10} \\
\bottomrule
\end{tabular}
}
\end{center}
\caption{\textbf{Ablations of the input image views.} "Front" means use only front-view image as input. "Surround" denotes using surround-view multiply images as input.}
\label{tab:view}
\end{table}

\begin{table}[]
\begin{center}
\centering
\renewcommand{\tabcolsep}{3.0pt}
\resizebox{0.98\linewidth}{!}{
\begin{tabular}{c|c|ccc|cccc}
\toprule
Image & \multirow{2}{*}{Acc. $(\%)$ $\uparrow$} & \multicolumn{3}{c|}{Path $($F1$)$ $\uparrow$} & \multicolumn{4}{c}{Speed $($F1$)$ $\uparrow$} \\
Tokens & & stra. & left & right & keep & acc. & dec. & stop  \\
\midrule
32$\times$6 & 69.36 & 95.41 & 87.71 & \textbf{91.18} & 79.46 & 55.05 & 49.55 & 80.63 \\
64$\times$6 & 70.42 & 95.41 & 88.01 & 90.78 & 80.07 & 55.97 & 51.37 & \textbf{81.52} \\
128$\times$6 & \textbf{71.21} & \textbf{95.60} & \textbf{89.37} & 90.09 & \textbf{80.18} & \textbf{58.83} & \textbf{61.99} & 80.10 \\
256$\times$6 & 56.13 & 94.45 & 88.70 & 88.59 & 69.52 & 39.40 & 48.27 & 40.68 \\
512$\times$6 & 0.00 & 0.00 & 0.00 & 0.00 & 0.00 & 0.00 & 0.00 & 0.00 \\
576$\times$6 & 0.00 & 0.00 & 0.00 & 0.00 & 0.00 & 0.00 & 0.00 & 0.00 \\
\bottomrule
\end{tabular}
}
\end{center}
\caption{\textbf{Ablations of the number of image tokens.} The term "Image Token" here refers to the tokens generated after each image is encoded and compressed by Senna's Driving Vision Adapter module.}
\label{tab:token}
\end{table}

\subsection{Training Strategy}
\label{sec:train}
We propose a three-stage training strategy for Senna-VLM. The first stage is Mix Pre-training, where we train the Driving Vision Adapter using single-image data while keeping the parameters of other modules in Senna-VLM frozen. This enables the mapping of image features to the LLM feature space. Mix refers to using data from multiple sources, including the instruction-following data used in LLaVA~\cite{liu2024llava} and our proposed driving scene description data. The second stage is Driving Fine-tuning, where we fine-tune Senna-VLM based on the planning-oriented QAs proposed in Sec.~\ref{sec:qa}, excluding the meta-action planning QAs. In this stage, surround-view multi-image inputs are used instead of single-image inputs. The third stage is Planning Fine-tuning, where we further fine-tune Senna-VLM using only the meta-action planning QAs. In both the second and third stages, we fine-tune all parameters of Senna-VLM, except for the Vision Encoder, which remains frozen.

For Senna-E2E, we use ground truth meta-actions as inputs in the training phase, while in the inference phase, it relies on the meta-actions predicted by Senna-VLM.

\section{Experiments}

\subsection{Experimental Settings}

\boldparagraph{Datasets.}
We evaluate the performance of the Senna on two datasets. The first is the nuScenes dataset~\cite{caesar2020nuscenes}, which consists of 1,000 driving scenes, and each scene lasts for about 20 seconds. It supports a variety of sensor data such as surround-view camera images, LiDAR point clouds, \etc. Besides, it also provides rich annotations, including 3D bounding boxes, ego vehicle trajectories, and road maps.

\begin{figure}[ht]
\centering
\includegraphics[width=0.98\linewidth]{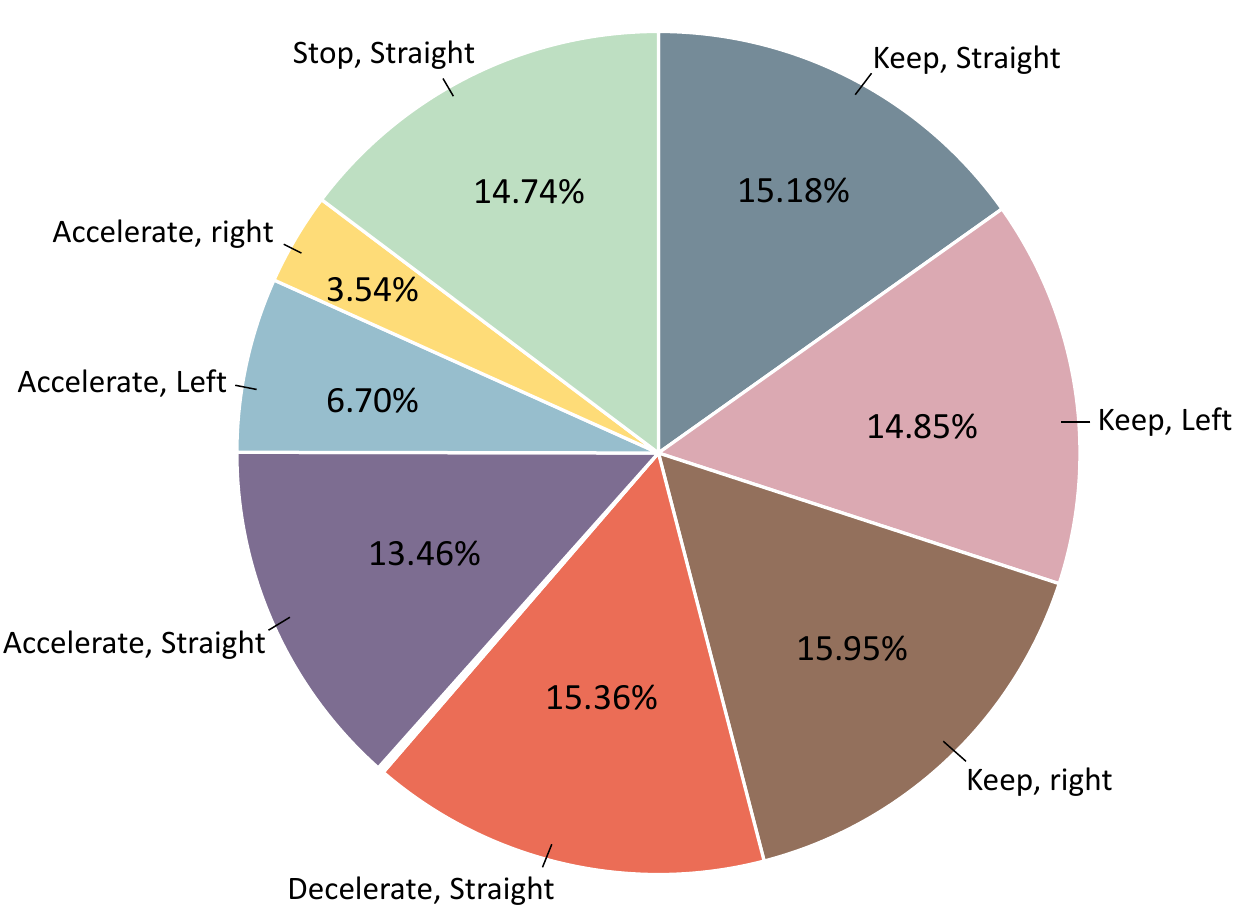} 
\caption{Visualization of Meta-action data distribution in the DriveX dataset.}
\label{fig:data}
\end{figure}

In addition to the nuScenes dataset, we also evaluate Senna on a large-scale dataset, DriveX, which consists of 1000K driving clips, each lasting 3 seconds. The DriveX dataset contains diverse and balanced driving data. Fig.~\ref{fig:data} illustrates the distribution of meta-actions within the dataset. We randomly sampled 800k clips for the training set and 200k for the validation set. Similar to nuScenes, DriveX provides surround-view images and diverse annotations, featuring a wider range of scenes and driving behaviors. Unless otherwise specified, Senna is trained on the DriveX training set, and the reported metrics are based on its validation set.

\begin{table}[]
\begin{center}
\centering
\renewcommand{\tabcolsep}{3.0pt}
\resizebox{0.98\linewidth}{!}{
\begin{tabular}{c|c|ccc|cccc}
\toprule
Train. & \multirow{2}{*}{Acc. $(\%)$ $\uparrow$} & \multicolumn{3}{c|}{Path $($F1$)$ $\uparrow$} & \multicolumn{4}{c}{Speed $($F1$)$ $\uparrow$} \\
Data & & stra. & left & right & keep & acc. & dec. & stop  \\
\midrule
100k & 57.45 & 94.47 & 89.69 & 85.78 & 71.38 & 38.82 & 42.18 & 66.89 \\
200k & 64.78 & 95.40 & 88.65 & 88.48 & 76.49 & 47.33 & 49.30 & 75.08 \\
400k & 67.09 & 95.40 & 89.97 & 89.22 & 77.62 & 54.49 & 51.04 & 76.76 \\
800k & \textbf{71.21} & \textbf{95.60} & \textbf{89.37} & \textbf{90.09} & \textbf{80.18} & \textbf{58.83} & \textbf{61.99} & \textbf{80.10} \\
\bottomrule
\end{tabular}
}
\end{center}
\caption{\textbf{Ablations of the training dataset size.}}
\label{tab:data}
\end{table}

\begin{table}[]
\begin{center}
\centering
\renewcommand{\tabcolsep}{2.0pt}
\resizebox{1.0\linewidth}{!}{
\begin{tabular}{l|ccccc}
\toprule
\multirow{2}{*}{Method} & Model & \# Input & \# Prompt  & \# Output & Decode \\
& Size & Image & Tokens & Tokens & Latency \\
\midrule
QwenVL~\cite{bai2023qwenvl} & 9.6B & 1 & 648 & 10 & 1.22s \\
VILA~\cite{lin2024vila} & 7.1B & 1 & 742 & 11 & 0.48s \\
LLaVA-1.5~\cite{liu2024llava1.5} & 7.1B & 1 & 808 & 11 & 0.51s \\
Senna$_{\rm 128}$ & 7.1B & 6 & 1000 & 11 & 0.58s \\
Senna$_{\rm 64}$ & 7.1B & 6 & 616 & 11 & 0.42s \\
Senna$_{\rm 32}$ & 7.1B & 6 & 424 & 11 & 0.35s \\
\bottomrule
\end{tabular}
}
\end{center}
\caption{\textbf{Ablation of the inference latency.} Speed is evaluated on an NVIDIA GeForce RTX 4090 GPU. Senna$_{\rm 128}$ denotes the Senna model encoding each image with 128 tokens, and similarly for Senna$_{\rm 64}$ and Senna$_{\rm 32}$.}
\label{tab:speed}
\end{table}

\boldparagraph{Metrics.}
Following previous works~\cite{drivemlm, tian2024drivevlm}, we use accuracy and F1 Score to evaluate high-level driving decisions. For driving scene descriptions, we adopt BLEU-4~\cite{papineni2002bleu}, CIDEr~\cite{vedantam2015cider}, and METEOR~\cite{banerjee2005meteor} as evaluation metrics. Finally, for trajectory planning, we employ displacement error and collision rate as the primary metrics.

\begin{table*}[]
\begin{center}
\centering
\renewcommand{\tabcolsep}{3.0pt}
\resizebox{0.95\textwidth}{!}{
\begin{tabular}{c|ccc|ccc|c|ccc|cccc}
\toprule
\multirow{2}{*}{ID} & \multicolumn{3}{c|}{Pre-training} & \multicolumn{3}{c|}{Fine-tuning} & \multirow{2}{*}{Acc. $(\%)$ $\uparrow$} & \multicolumn{3}{c|}{Path $($F1$)$ $\uparrow$} & \multicolumn{4}{c}{Speed $($F1$)$ $\uparrow$} \\
& Driving & General & Mix & Driving & Planning & Mix & & stra. & left & right & keep & acc. & dec. & stop  \\
\midrule
1 & \cmark & & &  &  & \cmark & 60.01 & 94.29 & 88.46 & 86.56 & 73.55 & 38.39 & 48.58 & 70.88 \\
2 & \cmark & & & \cmark & \cmark &  & 61.91 & 94.20 & 88.55 & 86.95 & 75.67 & 40.95 & 48.08 & 72.05 \\
3 & & \cmark &  &  &  & \cmark & 58.53 & 94.67 & 89.81 & 86.59 & 71.90 & 38.61 & 41.24 & 69.85 \\
4 & & \cmark &  & \cmark & \cmark &  & 65.12 & 95.29 & \textbf{90.12} & 88.46 & 75.78 & 51.38 & 52.25 & 77.22 \\
5 &  &  & \cmark &  &  & \cmark & 66.81 & 95.35 & 89.83 & 88.87 & 77.28 & 53.44 & 51.41 & 76.37 \\
6 &  &  & \cmark & \cmark & \cmark &  & \textbf{71.21} & \textbf{95.60} & 89.37 & \textbf{90.09} & \textbf{80.18} & \textbf{58.83} & \textbf{61.99} & \textbf{80.10} \\
\bottomrule
\end{tabular}
}
\end{center}
\caption{\textbf{Ablations of training pipelines.} In both the pre-training and fine-tuning phases, "Driving" refers to our proposed planning-oriented QAs, excluding the meta-action planning QAs. "Mix" refers to using both datasets during the respective phase. "General" represents the general-purpose dataset introduced by LLaVA~\cite{liu2024llava}. During the fine-tuning phase, "Planning" refers to the use of meta-action planning QA data. }
\label{tab:train}
\end{table*}

\subsection{Main Results}
\boldparagraph{Meta-action Planning.}
Tab.~\ref{tab:main} presents Senna’s performance in high-level planning and scene description, compared with state-of-the-art open-source LVLMs, including QwenVL, LLaVA, and VILA. The results for the first three rows are obtained by directly evaluating the original models. It can be seen that models using pre-trained weights perform poorly on driving tasks, as their training objectives are geared towards general understanding and conversation rather than being specifically tailored for driving-related tasks. To further validate Senna’s advantages, we also fine-tuned these models on the DriveX dataset using the same training pipeline. As shown, Senna outperforms the others in both high-level planning and scene description. Compared to the best results from other methods, Senna improves planning accuracy by 10.44\%. Furthermore, in the most critical driving safety decisions, such as deceleration, the F1 score increases from 52.68 to 61.99, achieving a 17.67\% improvement. This highlights Senna's superior ability in driving scene analysis and spatial understanding.

\boldparagraph{Trajectory Planning.}
we present Senna's trajectory planning performance on the nuScenes dataset in Tab.~\ref{tab:traj_nusc}. We replace VADv2 with VAD as the end-to-end model for a fair comparison. Compared to the previous SOTA method~\cite{tian2024drivevlm} that combines LVLMs with end-to-end models, Senna effectively reduces the average planning displacement error by 29.03\% and the collision rate by 20.00\%. To avoid the potential issues associated with using ego status features, as mentioned in previous work~\cite{li2024ego}, we also report results without utilizing ego status features. By initializing the model with pre-trained weights from the DriveX dataset and fine-tuning on the nuScenes dataset, Senna achieves state-of-the-art planning performance, significantly reducing the average planning displacement error by 40.28\% and decreasing the average collision rate by 45.45\% compared to VAD.~\cite{vad}. Senna's performance is significantly enhanced by pre-training on the DriveX dataset followed by fine-tuning on the nuScenes dataset, demonstrating its strong generalization and transferability.

Tab~\ref{tab:traj_drivex} presents the trajectory planning results on the DriveX dataset. In addition to the end-to-end model VADv2, we introduce two additional comparison models. The first model incorporates ground truth planning meta-actions as extra input features, aimed at verifying the performance upper bound of our proposed structured planning strategy. The second model, our reproduced DriveVLM~\cite{tian2024drivevlm}, predicts low-frequency trajectories instead of meta-actions, serving as the connector between the LVLM and the end-to-end model.

As shown in Tab.~\ref{tab:traj_drivex}, VADv2 which utilizes ground truth planning meta-actions, achieves the lowest planning error, validating the effectiveness of our proposed structured planning strategy. DriveVLM$\dagger$, which predicts low-frequency trajectories as connectors, shows only a marginal improvement over VADv2. In contrast, our proposed Senna delivers the best planning performance among all the method, which greatly reduces the average planning displacement error by 14.27\%.

\subsection{Ablation Study}
\vspace{-2mm}
\boldparagraph{Planning-oriented QAs.}
Tab.~\ref{tab:qa} presents the results of an ablation study validating the effectiveness of our proposed planning-oriented QAs. As Meta Action QA is a key component in our evaluation, Senna's planning accuracy is significantly impacted when it is excluded (ID 1). The absence of scenario description and traffic light detection QAs also affects acceleration and deceleration decisions, as starting and stopping in traffic light scenarios are closely tied to these QAs (ID 2-3). Moreover, omitting any other QAs leads to a decline in decision accuracy (ID 4-6). When all types of QAs are used to train Senna, it achieves optimal planning performance (ID 7).

\boldparagraph{Image View.}
As shown in Tab.~\ref{tab:view}, after incorporating surround-view multi-image input, Senna's planning accuracy increased from 64.91\% to 71.21\%. This demonstrates the effectiveness of Senna's image encoding strategy and the design of the surround-view prompt, which enhances the model's ability to understand the driving environment and make accurate planning decisions.

\boldparagraph{Number of Image Tokens.}
Tab.~\ref{tab:token} presents the results of using different numbers of image tokens when applying surround-view multi-image input. The original output of Senna's vision encoder contains 576 image tokens, as shown in the last row of the table. As observed, this leads to a model collapse, resulting in decoding failure. Reducing the number of image tokens to 512 does not resolve the issue. The model begins to output normally when the number of image tokens is reduced to 256, although the planning accuracy is still sub-optimal. The best results are achieved when the image token number is set to 128. Further reducing the image tokens to 64 and 32 has only a minor impact on planning accuracy, demonstrating the effectiveness and robustness of our proposed multi-image encoding strategy.

\boldparagraph{Training Dataset Size.}
We conducted an ablation study on the training dataset size, with the results presented in Tab.~\ref{tab:data}. As shown, when the number of training data is relatively small, Senna does not experience significant model degradation. However, as the training data increases, Senna's decision accuracy improves steadily, highlighting its excellent scaling capability.

\boldparagraph{Inference Speed.}
Tab.~\ref{tab:speed} compares the inference speed of Senna with other models, all evaluated using prompts of the same length. With six input images, compared to a single image for other models, Senna achieves comparable or even faster inference speeds due to its efficient multi-image encoding strategy. When encoding each image into 128 tokens, Senna’s decode latency is close to that of LLaVA-1.5~\cite{liu2024llava1.5}, despite handling more prompt tokens. Additionally, by reducing the number of tokens per image to 64 and 32, Senna further reduces its decode latency to 0.42 seconds and 0.35 seconds, respectively, representing a significant 17.6\% and 39.7\% improvement in inference speed over LLaVA-1.5.

\boldparagraph{Training Pipeline.}
We validate the effectiveness of the training pipeline, and the results are presented in Tab.~\ref{tab:train}. Senna achieves the best performance by applying our proposed three-stage training strategy: Mix Pre-training, Driving Fine-tuning, and Planning Fine-tuning.

\begin{figure*}[thbp]
\centering
\includegraphics[width=0.96\textwidth]{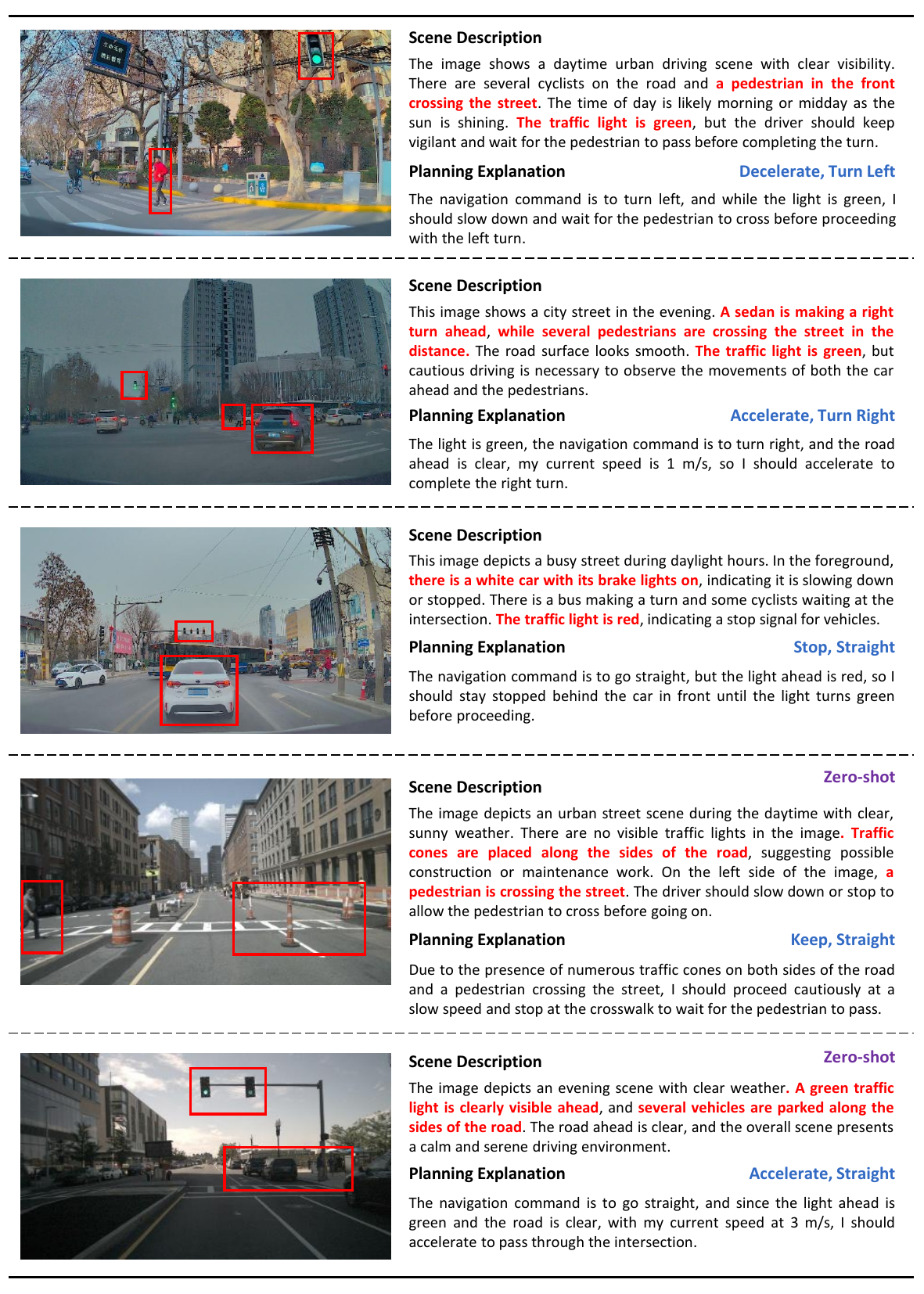}
\vspace{-8pt}
\caption{\textbf{Qualitative results of Senna.} The red boxes and text highlights key information that is relevant to driving decisions.}
\label{fig:vis}
\vspace{-25pt}
\end{figure*}

\subsection{Qualitative Results}
Fig.~\ref{fig:vis} presents qualitative results of Senna, showcasing its performance in driving scene description, high-level decision-making, and planning explanation on the DriveX validation set, as well as its zero-shot capabilities on the nuScenes dataset. Senna demonstrates impressive generalization abilities. In terms of scene description, Senna accurately captures key dynamic and static elements in driving scenarios, such as pedestrians crossing the road, traffic lights, and traffic cones. For planning explanation, based on precise scene descriptions and analysis, along with planning meta-actions, Senna effectively explains the reasoning behind its decisions. These detailed descriptions and planning explanations, in return, further enhance Senna's ability to make more informed decisions.

\section{Limitation}
This paper primarily validates the design and training strategies for a driving-oriented LVLM and its effective integration with an end-to-end model. However, during deployment, the inference speed of large models may hinder the system from meeting the real-time requirements of autonomous driving. This issue could potentially be mitigated in the future by utilizing smaller vision-language models, such as a 2B model, to reduce inference latency, combined with hardware-level optimizations to accelerate inference. Additionally, our experiments show that Senna's planning performance improves significantly as the dataset size increases, without any signs of diminishing returns. Given the limited data, we believe that Senna has not yet reached its performance ceiling, and there is still room for improvement as the training data continues to grow.


\section{Conclusion}
We propose Senna, an autonomous driving system that integrates an LVLM with an end-to-end model for structured planning, from high-level decision-making to low-level trajectory planning. Extensive experiments demonstrate Senna's superior performance, impressive cross-scenario generalization, and transferability, highlighting the potential of combining LVLMs with end-to-end models through language-based planning. Exploring non-predefined, granular language instructions to achieve controlled trajectory planning will be our future research direction.

\section*{Acknowledgments}
This work was supported by the National Natural Science Foundation of China (NSFC No. 62276108).


\bibliographystyle{IEEEtran}
\bibliography{egbib}

\begin{thebibliography}{10}
\providecommand{\url}[1]{#1}
\csname url@samestyle\endcsname
\providecommand{\newblock}{\relax}
\providecommand{\bibinfo}[2]{#2}
\providecommand{\BIBentrySTDinterwordspacing}{\spaceskip=0pt\relax}
\providecommand{\BIBentryALTinterwordstretchfactor}{4}
\providecommand{\BIBentryALTinterwordspacing}{\spaceskip=\fontdimen2\font plus
\BIBentryALTinterwordstretchfactor\fontdimen3\font minus \fontdimen4\font\relax}
\providecommand{\BIBforeignlanguage}[2]{{%
\expandafter\ifx\csname l@#1\endcsname\relax
\typeout{** WARNING: IEEEtran.bst: No hyphenation pattern has been}%
\typeout{** loaded for the language `#1'. Using the pattern for}%
\typeout{** the default language instead.}%
\else
\language=\csname l@#1\endcsname
\fi
#2}}
\providecommand{\BIBdecl}{\relax}
\BIBdecl

\bibitem{wang2022detr3d}
Y.~Wang, V.~C. Guizilini, T.~Zhang, Y.~Wang, H.~Zhao, and J.~Solomon, ``Detr3d: 3d object detection from multi-view images via 3d-to-2d queries,'' in \emph{CoRL}, 2022.

\bibitem{uniad}
Y.~Hu, J.~Yang, L.~Chen, K.~Li, C.~Sima, X.~Zhu, S.~Chai, S.~Du, T.~Lin, W.~Wang \emph{et~al.}, ``Planning-oriented autonomous driving,'' in \emph{CVPR}, 2023.

\bibitem{vad}
B.~Jiang, S.~Chen, Q.~Xu, B.~Liao, J.~Chen, H.~Zhou, Q.~Zhang, W.~Liu, C.~Huang, and X.~Wang, ``Vad: Vectorized scene representation for efficient autonomous driving,'' in \emph{ICCV}, 2023.

\bibitem{philion2020lss}
J.~Philion and S.~Fidler, ``Lift, splat, shoot: Encoding images from arbitrary camera rigs by implicitly unprojecting to 3d,'' in \emph{ECCV}, 2020.

\bibitem{li2022bevformer}
Z.~Li, W.~Wang, H.~Li, E.~Xie, C.~Sima, T.~Lu, Q.~Yu, and J.~Dai, ``Bevformer: Learning bird's-eye-view representation from multi-camera images via spatiotemporal transformers,'' \emph{arXiv preprint arXiv:2203.17270}, 2022.

\bibitem{liao2022maptr}
B.~Liao, S.~Chen, X.~Wang, T.~Cheng, Q.~Zhang, W.~Liu, and C.~Huang, ``Maptr: Structured modeling and learning for online vectorized hd map construction,'' \emph{arXiv preprint arXiv:2208.14437}, 2022.

\bibitem{chai2019multipath}
Y.~Chai, B.~Sapp, M.~Bansal, and D.~Anguelov, ``Multipath: Multiple probabilistic anchor trajectory hypotheses for behavior prediction,'' \emph{arXiv preprint arXiv:1910.05449}, 2019.

\bibitem{gu2022vip3d}
J.~Gu, C.~Hu, T.~Zhang, X.~Chen, Y.~Wang, Y.~Wang, and H.~Zhao, ``Vip3d: End-to-end visual trajectory prediction via 3d agent queries,'' \emph{arXiv preprint arXiv:2208.01582}, 2022.

\bibitem{jiang2022pip}
B.~Jiang, S.~Chen, X.~Wang, B.~Liao, T.~Cheng, J.~Chen, H.~Zhou, Q.~Zhang, W.~Liu, and C.~Huang, ``Perceive, interact, predict: Learning dynamic and static clues for end-to-end motion prediction,'' \emph{arXiv preprint arXiv:2212.02181}, 2022.

\bibitem{toromanoff2020end}
M.~Toromanoff, E.~Wirbel, and F.~Moutarde, ``End-to-end model-free reinforcement learning for urban driving using implicit affordances,'' in \emph{CVPR}, 2020.

\bibitem{transfuser}
A.~Prakash, K.~Chitta, and A.~Geiger, ``Multi-modal fusion transformer for end-to-end autonomous driving,'' in \emph{CVPR}, 2021.

\bibitem{hu2022stp3}
S.~Hu, L.~Chen, P.~Wu, H.~Li, J.~Yan, and D.~Tao, ``St-p3: End-to-end vision-based autonomous driving via spatial-temporal feature learning,'' in \emph{ECCV}, 2022.

\bibitem{liu2024llava}
H.~Liu, C.~Li, Q.~Wu, and Y.~J. Lee, ``Visual instruction tuning,'' in \emph{NeurIPS}, 2024.

\bibitem{bai2023qwenvl}
J.~Bai, S.~Bai, S.~Yang, S.~Wang, S.~Tan, P.~Wang, J.~Lin, C.~Zhou, and J.~Zhou, ``Qwen-vl: A frontier large vision-language model with versatile abilities,'' \emph{arXiv preprint arXiv:2308.12966}, 2023.

\bibitem{wang2023cogvlm}
W.~Wang, Q.~Lv, W.~Yu, W.~Hong, J.~Qi, Y.~Wang, J.~Ji, Z.~Yang, L.~Zhao, X.~Song \emph{et~al.}, ``Cogvlm: Visual expert for pretrained language models,'' \emph{arXiv preprint arXiv:2311.03079}, 2023.

\bibitem{chen2024internvl}
Z.~Chen, J.~Wu, W.~Wang, W.~Su, G.~Chen, S.~Xing, M.~Zhong, Q.~Zhang, X.~Zhu, L.~Lu \emph{et~al.}, ``Internvl: Scaling up vision foundation models and aligning for generic visual-linguistic tasks,'' in \emph{CVPR}, 2024.

\bibitem{achiam2023gpt4}
J.~Achiam, S.~Adler, S.~Agarwal, L.~Ahmad, I.~Akkaya, F.~L. Aleman, D.~Almeida, J.~Altenschmidt, S.~Altman, S.~Anadkat \emph{et~al.}, ``Gpt-4 technical report,'' \emph{arXiv preprint arXiv:2303.08774}, 2023.

\bibitem{botvinick2008hierarchical}
M.~M. Botvinick, ``Hierarchical models of behavior and prefrontal function,'' \emph{Trends in cognitive sciences}, 2008.

\bibitem{koechlin2003cognitive}
E.~Koechlin, C.~Ody, and F.~Kouneiher, ``The architecture of cognitive control in the human prefrontal cortex,'' \emph{Science}, 2003.

\bibitem{badre2008cognitive}
D.~Badre, ``Cognitive control, hierarchy, and the rostro--caudal organization of the frontal lobes,'' \emph{Trends in cognitive sciences}, 2008.

\bibitem{drivegpt4}
Z.~Xu, Y.~Zhang, E.~Xie, Z.~Zhao, Y.~Guo, K.~K. Wong, Z.~Li, and H.~Zhao, ``Drivegpt4: Interpretable end-to-end autonomous driving via large language model,'' \emph{arXiv preprint arXiv:2310.01412}, 2023.

\bibitem{driving-with-llms}
L.~Chen, O.~Sinavski, J.~H{\"u}nermann, A.~Karnsund, A.~J. Willmott, D.~Birch, D.~Maund, and J.~Shotton, ``Driving with llms: Fusing object-level vector modality for explainable autonomous driving,'' \emph{arXiv preprint arXiv:2310.01957}, 2023.

\bibitem{drivemlm}
W.~Wang, J.~Xie, C.~Hu, H.~Zou, J.~Fan, W.~Tong, Y.~Wen, S.~Wu, H.~Deng, Z.~Li \emph{et~al.}, ``Drivemlm: Aligning multi-modal large language models with behavioral planning states for autonomous driving,'' \emph{arXiv preprint arXiv:2312.09245}, 2023.

\bibitem{wang2024omnidrive}
S.~Wang, Z.~Yu, X.~Jiang, S.~Lan, M.~Shi, N.~Chang, J.~Kautz, Y.~Li, and J.~M. Alvarez, ``Omnidrive: A holistic llm-agent framework for autonomous driving with 3d perception, reasoning and planning,'' \emph{arXiv preprint arXiv:2405.01533}, 2024.

\bibitem{frieder2024mathematical}
S.~Frieder, L.~Pinchetti, R.-R. Griffiths, T.~Salvatori, T.~Lukasiewicz, P.~Petersen, and J.~Berner, ``Mathematical capabilities of chatgpt,'' in \emph{NeurIPS}, 2024.

\bibitem{hendrycks2021measuring}
D.~Hendrycks, C.~Burns, S.~Kadavath, A.~Arora, S.~Basart, E.~Tang, D.~Song, and J.~Steinhardt, ``Measuring mathematical problem solving with the math dataset,'' \emph{arXiv preprint arXiv:2103.03874}, 2021.

\bibitem{tian2024drivevlm}
X.~Tian, J.~Gu, B.~Li, Y.~Liu, C.~Hu, Y.~Wang, K.~Zhan, P.~Jia, X.~Lang, and H.~Zhao, ``Drivevlm: The convergence of autonomous driving and large vision-language models,'' \emph{arXiv preprint arXiv:2402.12289}, 2024.

\bibitem{lin2024vila}
J.~Lin, H.~Yin, W.~Ping, P.~Molchanov, M.~Shoeybi, and S.~Han, ``Vila: On pre-training for visual language models,'' in \emph{CVPR}, 2024.

\bibitem{zhou2024elm}
Y.~Zhou, L.~Huang, Q.~Bu, J.~Zeng, T.~Li, H.~Qiu, H.~Zhu, M.~Guo, Y.~Qiao, and H.~Li, ``Embodied understanding of driving scenarios,'' \emph{arXiv preprint arXiv:2403.04593}, 2024.

\bibitem{liu2024llava1.5}
H.~Liu, C.~Li, Y.~Li, and Y.~J. Lee, ``Improved baselines with visual instruction tuning,'' in \emph{CVPR}, 2024.

\bibitem{sima2023drivelm}
C.~Sima, K.~Renz, K.~Chitta, L.~Chen, H.~Zhang, C.~Xie, P.~Luo, A.~Geiger, and H.~Li, ``Drivelm: Driving with graph visual question answering,'' \emph{arXiv preprint arXiv:2312.14150}, 2023.

\bibitem{wu2023nuprompt}
D.~Wu, W.~Han, T.~Wang, Y.~Liu, X.~Zhang, and J.~Shen, ``Language prompt for autonomous driving,'' \emph{arXiv preprint arXiv:2309.04379}, 2023.

\bibitem{qian2024nuscenes-qa}
T.~Qian, J.~Chen, L.~Zhuo, Y.~Jiao, and Y.-G. Jiang, ``Nuscenes-qa: A multi-modal visual question answering benchmark for autonomous driving scenario,'' in \emph{AAAI}, 2024.

\bibitem{caesar2020nuscenes}
H.~Caesar, V.~Bankiti, A.~H. Lang, S.~Vora, V.~E. Liong, Q.~Xu, A.~Krishnan, Y.~Pan, G.~Baldan, and O.~Beijbom, ``nuscenes: A multimodal dataset for autonomous driving,'' in \emph{CVPR}, 2020.

\bibitem{paden2016survey}
B.~Paden, M.~{\v{C}}{\'a}p, S.~Z. Yong, D.~Yershov, and E.~Frazzoli, ``A survey of motion planning and control techniques for self-driving urban vehicles,'' \emph{IEEE Transactions on intelligent vehicles}, 2016.

\bibitem{thrun2006stanley}
S.~Thrun, M.~Montemerlo, H.~Dahlkamp, D.~Stavens, A.~Aron, J.~Diebel, P.~Fong, J.~Gale, M.~Halpenny, G.~Hoffmann \emph{et~al.}, ``Stanley: The robot that won the darpa grand challenge,'' \emph{Journal of field Robotics}, 2006.

\bibitem{urmson2008autonomous}
C.~Urmson, J.~Anhalt, D.~Bagnell, C.~Baker, R.~Bittner, M.~Clark, J.~Dolan, D.~Duggins, T.~Galatali, C.~Geyer \emph{et~al.}, ``Autonomous driving in urban environments: Boss and the urban challenge,'' \emph{Journal of field Robotics}, 2008.

\bibitem{codevilla2019exploring}
F.~Codevilla, E.~Santana, A.~M. L{\'o}pez, and A.~Gaidon, ``Exploring the limitations of behavior cloning for autonomous driving,'' in \emph{ICCV}, 2019.

\bibitem{pomerleau1988alvinn}
D.~A. Pomerleau, ``Alvinn: An autonomous land vehicle in a neural network,'' in \emph{NeurIPS}, 1988.

\bibitem{roach}
Z.~Zhang, A.~Liniger, D.~Dai, F.~Yu, and L.~Van~Gool, ``End-to-end urban driving by imitating a reinforcement learning coach,'' in \emph{ICCV}, 2021.

\bibitem{driveadapter}
X.~Jia, Y.~Gao, L.~Chen, J.~Yan, P.~L. Liu, and H.~Li, ``Driveadapter: Breaking the coupling barrier of perception and planning in end-to-end autonomous driving,'' in \emph{ICCV}, 2023.

\bibitem{maptrv2}
B.~Liao, S.~Chen, Y.~Zhang, B.~Jiang, Q.~Zhang, W.~Liu, C.~Huang, and X.~Wang, ``Maptrv2: An end-to-end framework for online vectorized hd map construction,'' \emph{arXiv preprint arXiv:2308.05736}, 2023.

\bibitem{chen2024vadv2}
S.~Chen, B.~Jiang, H.~Gao, B.~Liao, Q.~Xu, Q.~Zhang, C.~Huang, W.~Liu, and X.~Wang, ``Vadv2: End-to-end vectorized autonomous driving via probabilistic planning,'' \emph{arXiv preprint arXiv:2402.13243}, 2024.

\bibitem{vaswani2017attention}
A.~Vaswani, N.~Shazeer, N.~Parmar, J.~Uszkoreit, L.~Jones, A.~N. Gomez, {\L}.~Kaiser, and I.~Polosukhin, ``Attention is all you need,'' in \emph{NeurIPS}, 2017.

\bibitem{gpt3}
T.~Brown, B.~Mann, N.~Ryder, M.~Subbiah, J.~D. Kaplan, P.~Dhariwal, A.~Neelakantan, P.~Shyam, G.~Sastry, A.~Askell \emph{et~al.}, ``Language models are few-shot learners,'' in \emph{NeurIPS}, 2020.

\bibitem{touvron2023llama}
H.~Touvron, T.~Lavril, G.~Izacard, X.~Martinet, M.-A. Lachaux, T.~Lacroix, B.~Rozi{\`e}re, N.~Goyal, E.~Hambro, F.~Azhar \emph{et~al.}, ``Llama: Open and efficient foundation language models,'' \emph{arXiv preprint arXiv:2302.13971}, 2023.

\bibitem{anil2023palm2}
R.~Anil, A.~M. Dai, O.~Firat, M.~Johnson, D.~Lepikhin, A.~Passos, S.~Shakeri, E.~Taropa, P.~Bailey, Z.~Chen \emph{et~al.}, ``Palm 2 technical report,'' \emph{arXiv preprint arXiv:2305.10403}, 2023.

\bibitem{yang2024qwen2}
A.~Yang, B.~Yang, B.~Hui, B.~Zheng, B.~Yu, C.~Zhou, C.~Li, C.~Li, D.~Liu, F.~Huang \emph{et~al.}, ``Qwen2 technical report,'' \emph{arXiv preprint arXiv:2407.10671}, 2024.

\bibitem{zhu2023minigpt4}
D.~Zhu, J.~Chen, X.~Shen, X.~Li, and M.~Elhoseiny, ``Minigpt-4: Enhancing vision-language understanding with advanced large language models,'' \emph{arXiv preprint arXiv:2304.10592}, 2023.

\bibitem{dai2023instructblip}
W.~Dai, J.~Li, D.~Li, A.~M.~H. Tiong, J.~Zhao, W.~Wang, B.~Li, P.~Fung, and S.~Hoi, ``Instructblip: Towards general-purpose vision-language models with instruction tuning,'' 2023.

\bibitem{radford2021clip}
A.~Radford, J.~W. Kim, C.~Hallacy, A.~Ramesh, G.~Goh, S.~Agarwal, G.~Sastry, A.~Askell, P.~Mishkin, J.~Clark \emph{et~al.}, ``Learning transferable visual models from natural language supervision,'' in \emph{ICML}, 2021.

\bibitem{fang2023eva}
Y.~Fang, W.~Wang, B.~Xie, Q.~Sun, L.~Wu, X.~Wang, T.~Huang, X.~Wang, and Y.~Cao, ``Eva: Exploring the limits of masked visual representation learning at scale,'' in \emph{CVPR}, 2023.

\bibitem{li2022blip}
J.~Li, D.~Li, C.~Xiong, and S.~Hoi, ``Blip: Bootstrapping language-image pre-training for unified vision-language understanding and generation,'' in \emph{ICML}, 2022.

\bibitem{li2023blip2}
J.~Li, D.~Li, S.~Savarese, and S.~Hoi, ``Blip-2: Bootstrapping language-image pre-training with frozen image encoders and large language models,'' in \emph{ICML}, 2023.

\bibitem{jia2021align}
C.~Jia, Y.~Yang, Y.~Xia, Y.-T. Chen, Z.~Parekh, H.~Pham, Q.~Le, Y.-H. Sung, Z.~Li, and T.~Duerig, ``Scaling up visual and vision-language representation learning with noisy text supervision,'' in \emph{ICML}, 2021.

\bibitem{lu2019vilbert}
J.~Lu, D.~Batra, D.~Parikh, and S.~Lee, ``Vilbert: Pretraining task-agnostic visiolinguistic representations for vision-and-language tasks,'' in \emph{NeurIPS}, 2019.

\bibitem{tan2019lxmert}
H.~Tan and M.~Bansal, ``Lxmert: Learning cross-modality encoder representations from transformers,'' \emph{arXiv preprint arXiv:1908.07490}, 2019.

\bibitem{wang2024qwen2vl}
P.~Wang, S.~Bai, S.~Tan, S.~Wang, Z.~Fan, J.~Bai, K.~Chen, X.~Liu, J.~Wang, W.~Ge \emph{et~al.}, ``Qwen2-vl: Enhancing vision-language model's perception of the world at any resolution,'' \emph{arXiv preprint arXiv:2409.12191}, 2024.

\bibitem{languagempc}
H.~Sha, Y.~Mu, Y.~Jiang, L.~Chen, C.~Xu, P.~Luo, S.~E. Li, M.~Tomizuka, W.~Zhan, and M.~Ding, ``Languagempc: Large language models as decision makers for autonomous driving,'' \emph{arXiv preprint arXiv:2310.03026}, 2023.

\bibitem{dosovitskiy2017carla}
A.~Dosovitskiy, G.~Ros, F.~Codevilla, A.~Lopez, and V.~Koltun, ``Carla: An open urban driving simulator,'' in \emph{CoRL}, 2017.

\bibitem{wu2023refer-kitti}
D.~Wu, W.~Han, T.~Wang, X.~Dong, X.~Zhang, and J.~Shen, ``Referring multi-object tracking,'' in \emph{CVPR}, 2023.

\bibitem{deruyttere2019talk2car}
T.~Deruyttere, S.~Vandenhende, D.~Grujicic, L.~Van~Gool, and M.-F. Moens, ``Talk2car: Taking control of your self-driving car,'' \emph{arXiv preprint arXiv:1909.10838}, 2019.

\bibitem{kim2018bdd-x}
J.~Kim, A.~Rohrbach, T.~Darrell, J.~Canny, and Z.~Akata, ``Textual explanations for self-driving vehicles,'' in \emph{Proceedings of the European conference on computer vision (ECCV)}, 2018, pp. 563--578.

\bibitem{zheng2023vicuna}
L.~Zheng, W.-L. Chiang, Y.~Sheng, S.~Zhuang, Z.~Wu, Y.~Zhuang, Z.~Lin, Z.~Li, D.~Li, E.~Xing \emph{et~al.}, ``Judging llm-as-a-judge with mt-bench and chatbot arena,'' in \emph{NeurIPS}, 2023.

\bibitem{hendrycks2016gelu}
D.~Hendrycks and K.~Gimpel, ``Gaussian error linear units (gelus),'' \emph{arXiv preprint arXiv:1606.08415}, 2016.

\bibitem{ratliff2006il}
N.~D. Ratliff, J.~A. Bagnell, and M.~A. Zinkevich, ``Maximum margin planning,'' in \emph{Proceedings of the 23rd international conference on Machine learning}, 2006, pp. 729--736.

\bibitem{zeng2019nmp}
W.~Zeng, W.~Luo, S.~Suo, A.~Sadat, B.~Yang, S.~Casas, and R.~Urtasun, ``End-to-end interpretable neural motion planner,'' in \emph{Proceedings of the IEEE/CVF Conference on Computer Vision and Pattern Recognition}, 2019, pp. 8660--8669.

\bibitem{hu2021ff}
P.~Hu, A.~Huang, J.~Dolan, D.~Held, and D.~Ramanan, ``Safe local motion planning with self-supervised freespace forecasting,'' in \emph{CVPR}, 2021.

\bibitem{khurana2022eo}
T.~Khurana, P.~Hu, A.~Dave, J.~Ziglar, D.~Held, and D.~Ramanan, ``Differentiable raycasting for self-supervised occupancy forecasting,'' in \emph{ECCV}, 2022.

\bibitem{papineni2002bleu}
K.~Papineni, S.~Roukos, T.~Ward, and W.-J. Zhu, ``Bleu: a method for automatic evaluation of machine translation,'' in \emph{ACL}, 2002.

\bibitem{vedantam2015cider}
R.~Vedantam, C.~Lawrence~Zitnick, and D.~Parikh, ``Cider: Consensus-based image description evaluation,'' in \emph{CVPR}, 2015.

\bibitem{banerjee2005meteor}
S.~Banerjee and A.~Lavie, ``Meteor: An automatic metric for mt evaluation with improved correlation with human judgments,'' in \emph{Proceedings of the acl workshop on intrinsic and extrinsic evaluation measures for machine translation and/or summarization}, 2005.

\bibitem{li2024ego}
Z.~Li, Z.~Yu, S.~Lan, J.~Li, J.~Kautz, T.~Lu, and J.~M. Alvarez, ``Is ego status all you need for open-loop end-to-end autonomous driving?'' in \emph{CVPR}, 2024.

\end{thebibliography}

\vspace{-4mm}
\begin{IEEEbiography}
[{\includegraphics[width=1in,height=1.25in,clip,keepaspectratio]{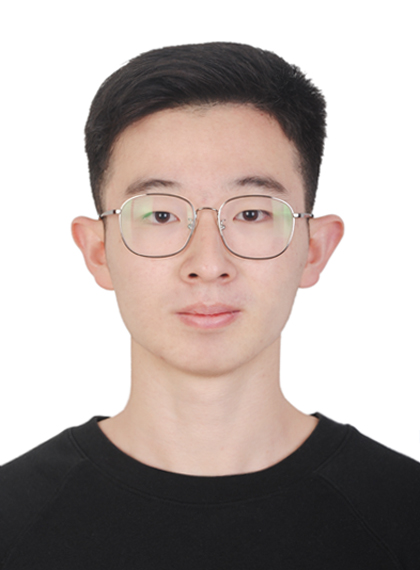}}]{Bo Jiang} received his B.E. degree in data science and big data technology from Central South University, China, in 2021. Currently, he is a Ph.D. candidate at Huazhong University of Science and Technology, China. His research interest lies in the field of multimodal learning, autonomous driving, and embodied AI.
\end{IEEEbiography}

\vspace{-4mm}
\begin{IEEEbiography}
[{\includegraphics[width=1in,height=1.25in,clip,keepaspectratio]{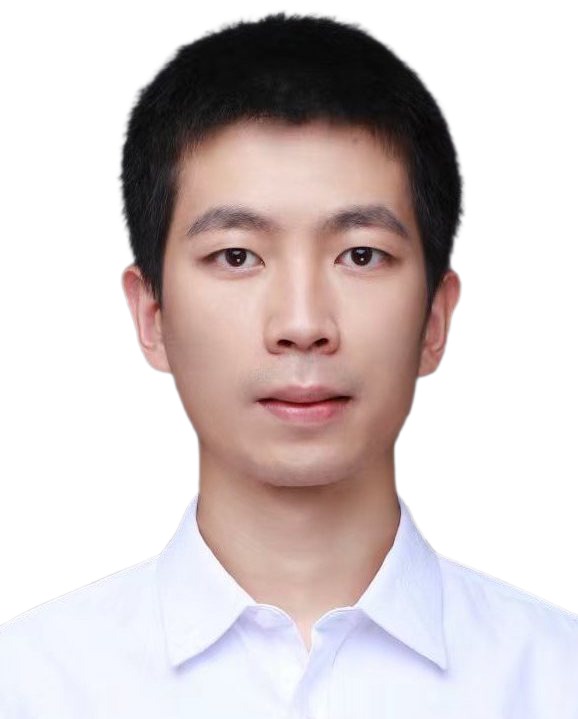}}]{Shaoyu Chen} received the B.E. degree from the School of Electronic Information and Communications, Huazhong University of Science and Technology (HUST), Wuhan, China, in 2019. He is currently pursuing a Ph.D. degree at HUST. His research interests include 3D vision and autonomous driving.
\end{IEEEbiography}

\vspace{-4mm}
\begin{IEEEbiography}[{\includegraphics[width=1in,height=1.25in,clip,keepaspectratio]{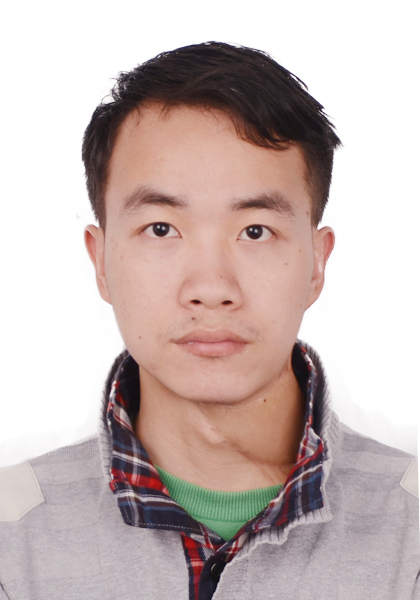}}]{Bencheng Liao} received the B.E. degree from School of Electronic Information and Communications, Huazhong University of Science and Technology, Wuhan, China, in 2020. He is currently a PhD candidate at the Institute of Artificial Intelligence and School of Electronic Information and Communications, Huazhong University of Science and Technology. His research interests include object detection, 3D vision, and autonomous driving.
\end{IEEEbiography}

\vspace{-4mm}
\begin{IEEEbiography}[{\includegraphics[width=1in,height=1.25in,clip,keepaspectratio]{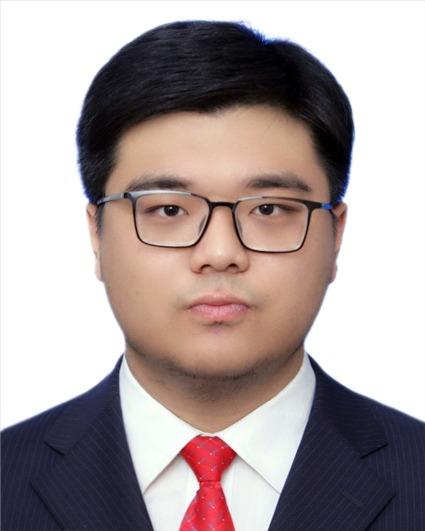}}]{Xingyu Zhang} received his Bachelor’s degree in Computer Science and Technology from Xidian University in 2020 and the Master’s degree in Computer Science and Technology from Xi’an Jiaotong University in 2023. He is currently a researcher at Horizon Robotics, focusing on autonomous driving, large multi-modal model, and world model. His research interests include autonomous driving with Vision-Language Models and world model simulations.
\end{IEEEbiography}

\vspace{-4mm}
\begin{IEEEbiography}[{\includegraphics[width=1in,height=1.25in,clip,keepaspectratio]{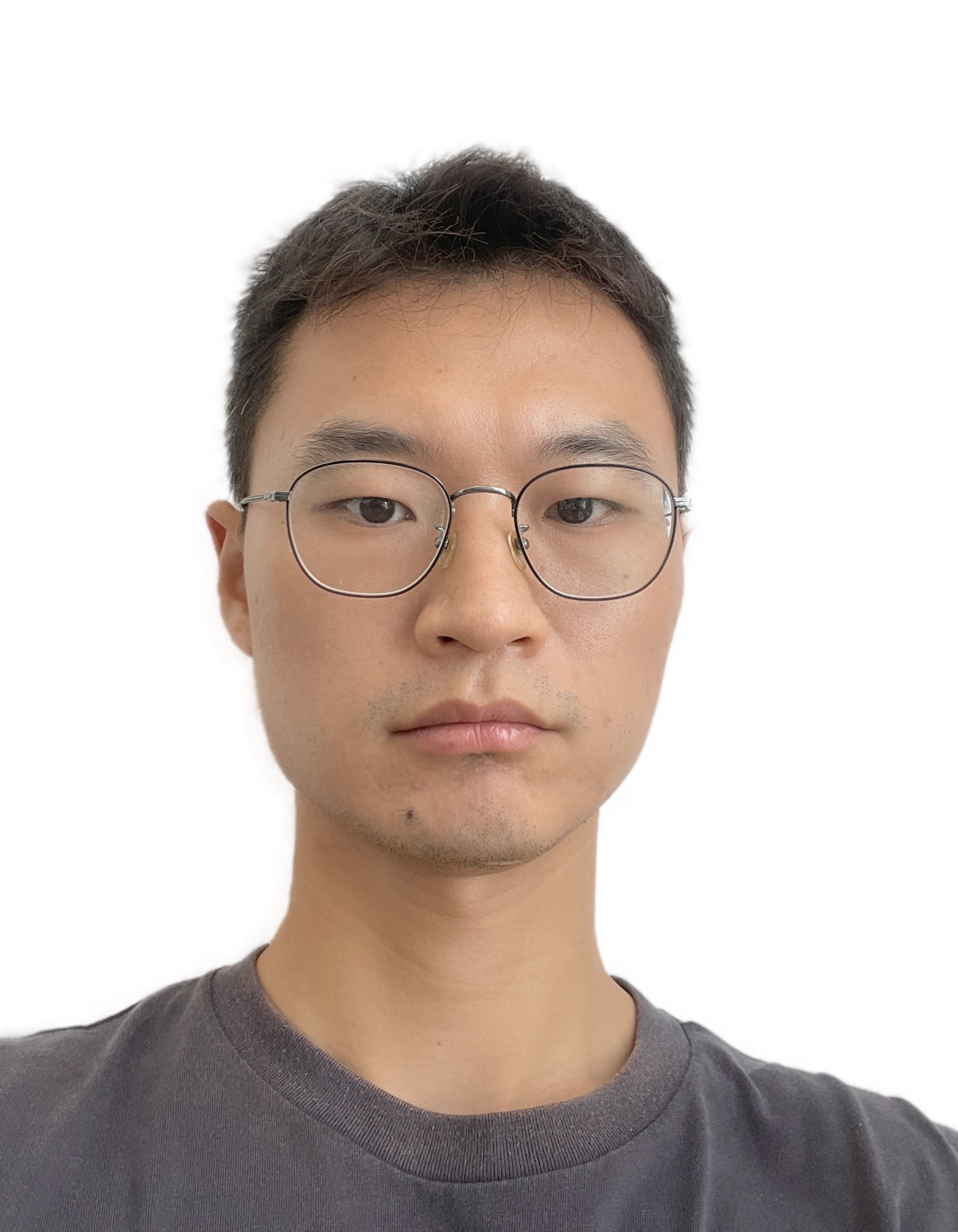}}]{Wei Yin} is a research scientist at Horizon Robotics. Before that, he received his Ph.D. degree at Australian Institute of Machine Learning (AIML), University of Adelaide. His research interests include the world model, 3D reconstruction, and depth estimation.
\end{IEEEbiography}

\vspace{-4mm}
\begin{IEEEbiography}[{\includegraphics[width=1in,height=1.25in,clip,keepaspectratio]{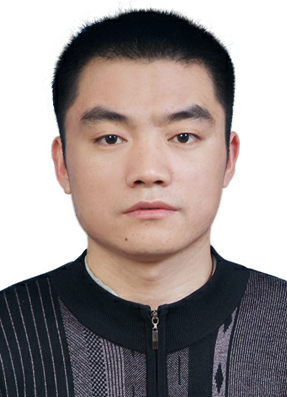}}]{Qian Zhang} received the B.E. and M.S. degrees from Central South University, Changsha, China, in 2008 and 2011, respectively, and the Ph.D. degree in pattern recognition and intelligent systems from the Institute of Automation, Chinese Academy of Sciences, Beijing, China, in 2014. His current research interests include computer vision and machine learning.
\end{IEEEbiography}

\vspace{-4mm}
\begin{IEEEbiography}[{\includegraphics[width=1in,height=1.25in,clip,keepaspectratio]{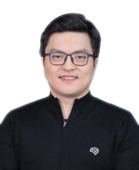}}]{Chang Huang} received the B.S. and Ph.D. degrees in Computer Science from Tsinghua University, in 2003 and 2007, respectively. He is the co-founder \& CTO of Horizon Robotics. He led the in-house research and development behind the company's BPU (Brain Processing Unit) computing processor architecture, computer vision, and deep learning.
\end{IEEEbiography}

\vspace{-4mm}
\begin{IEEEbiography}[{\includegraphics[width=1in,height=1.25in,clip,keepaspectratio]{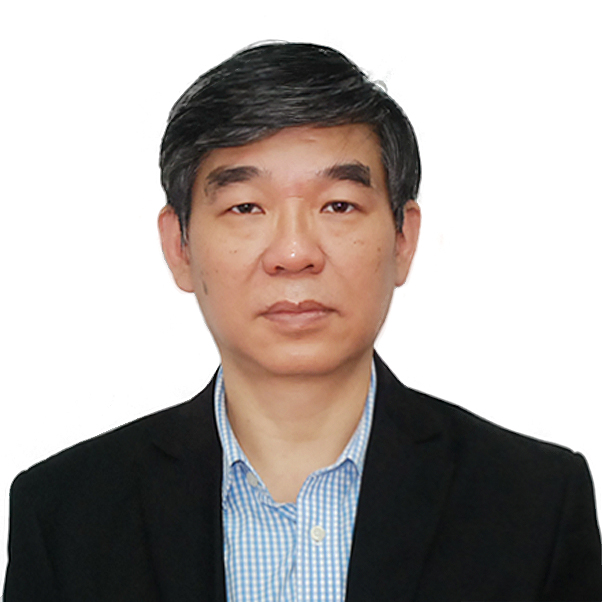}}]{Wenyu Liu (SM’15)} 
    received the B.S. degree in Computer Science from Tsinghua University, Beijing, China, in 1986, and the M.S. and Ph.D. degrees, both in Electronics and Information Engineering, from Huazhong University of Science and Technology (HUST), Wuhan, China, in 1991 and 2001, respectively. He is now a professor and associate dean of the School of Electronic Information and Communications, HUST. His research areas include computer vision, multimedia, and machine learning.
\end{IEEEbiography}

\vspace{-4mm}
\begin{IEEEbiography}[{\includegraphics[width=1in,height=1.25in,clip,keepaspectratio]{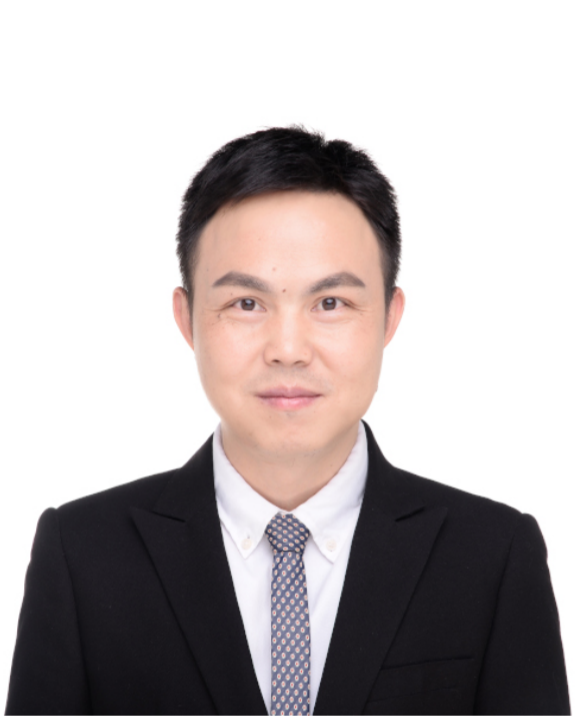}}]{Xinggang Wang} received the B.S. and Ph.D. degrees from Huazhong University of Science and Technology (HUST), Wuhan, China, in 2009 and 2014, respectively. He is now a professor at the School of Electronic Information and Communications, HUST. He serves as Co-Editor-in-Chief of Image and Vision Computing, associate editor of Pattern Recognition, and area chair of CVPR and ICCV. His research interests include computer vision and deep learning.
\end{IEEEbiography}

\end{document}